\documentclass[times, review, 10pt]{elsarticle}

\usepackage{amssymb}
\usepackage{framed,multirow}

\usepackage{latexsym}
\usepackage{textcomp,booktabs}
\usepackage{amssymb}
\usepackage{pifont}
\usepackage{url}
\usepackage{cite}
\usepackage{natbib}
\usepackage{xcolor}
\definecolor{newcolor}{rgb}{.8,.349,.1}
\usepackage{marvosym}

\usepackage{hyperref} 
\hypersetup{hidelinks, colorlinks=true, linkcolor=red, citecolor=blue, urlcolor=magenta}

\usepackage{amsthm}
\usepackage{amsmath}

\journal{Pattern Recognition}

\begin{document}

\begin{frontmatter}




\title{  
Visible-Light Imaging Diagnosis of Neutral Particle Emission Tomography in the Tokamak Divertor: An Efficient Transformer-based Surrogate Model
}






\author{Xiao Wang\textsuperscript{1}, Hao Si\textsuperscript{1}, Qiang Chen\textsuperscript{1}, Yu-Xiang Zhang\textsuperscript{1}, Beihe Zhang\textsuperscript{3},  
Jianhua Yang\textsuperscript{3,*}, Qingquan Yang\textsuperscript{3}, 
Dengdi Sun\textsuperscript{2,*}, 
Wanli Lyu\textsuperscript{1}, 
Guosheng Xu\textsuperscript{3}, 
Jin Tang\textsuperscript{1}}

\address{
1. School of Computer Science and Technology, Anhui University, Hefei 230601, China. \\
2. School of Artificial Intelligence, Anhui University, Hefei 230601, China. \\
3. Institute of Plasma Physics, Hefei Institutes of Physical Science, Chinese Academy of Sciences, Hefei 230031, China. \\
}







\begin{abstract}
Nuclear fusion has made significant progress in recent years and is expected to become one of the most important pathways to addressing global energy challenges. This paper focuses on observing plasma using visible-light cameras, analyzing its spatio-temporal motion cues, and predicting the {two-dimensional} spatial distribution of light intensity, aiming to provide a foundational basis for future scientific experiments using deep neural networks. Specifically, 
{we propose \textbf{Delta-InvFormer}, a novel backbone network centered on a differential Transformer.}
The key insight is that by taking consecutive video frames as input, we can better capture the dynamics of the plasma. Moreover, spatial and temporal differential self-attention effectively mitigates interference from noisy signals, ensuring high-quality feature extraction. These features are then fused into a compact and informative representation, which is fed into a decoder network to predict the distribution. Based on real experimental data collected from the {Experimental Advanced Superconducting Tokamak (EAST)} large-scale scientific facility, our results demonstrate that the proposed model not only significantly accelerates traditional methods for distribution prediction but also achieves {competitive reconstruction accuracy.}
The source code of this paper will be released upon acceptance. 
\end{abstract}

\end{frontmatter}


\section{Introduction}

Nuclear fusion has always been one of the core pathways for humanity's pursuit of clean energy. In recent years, significant breakthroughs have been achieved based on various fusion devices, such as EAST\footnote{\url{http://english.ipp.cas.cn/}}, JET, JT-60SA, and HL-3. Researchers in the field of nuclear fusion are also exploring the integration of artificial intelligence (AI) technologies to accelerate advancements in fusion research~\citep{wang2025xiHeFusion}. At the same time, the virtually limitless energy provided by fusion has the potential to address the high power consumption demands of AI development. Therefore, the synergistic convergence of these two fields holds great promise and is highly anticipated.

\begin{figure*}
\centering
\includegraphics[width=1\linewidth]{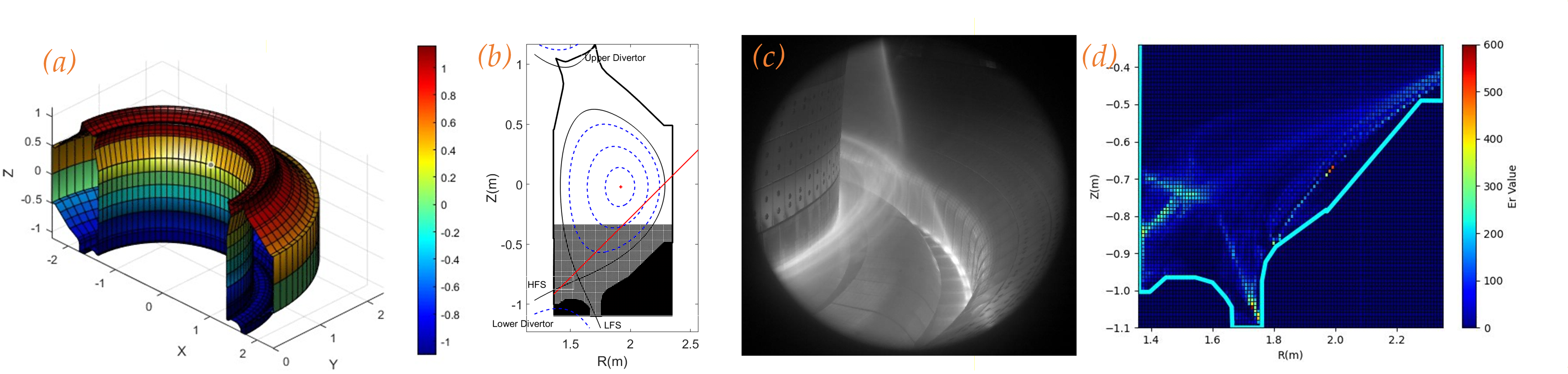}
\caption{An illustration of 
(a) 3D interface diagram of the Tokamak device; 
(b) EAST cross-sectional diagram and the area of the lower divertor;
(c) Visible light imaging from the observation window; 
(d) 2D planar image formed after inversion.}
\label{fig:firstIMG}
\end{figure*}

In this paper, we address the research problem of estimating the distribution of neutral particles in the divertor and Scrape-Off Layer (SOL) regions, a critical task for both fundamental plasma physics studies and the routine operation of fusion devices~\citep{zhang2025tomography}. From the perspective of artificial intelligence, this task involves using visible light cameras on the EAST device to observe the plasma, and then employing a machine learning model to predict the brightness distribution of each light ray in 3D space. Figure~\ref {fig:firstIMG} (a) is a 3D view of the EAST device, Figure~\ref{fig:firstIMG} (b) shows a schematic diagram of the EAST device's poloidal cross-section. Figure~\ref{fig:firstIMG} (c) is an image that is captured by a high-resolution, tangentially viewing visible camera system on the EAST device, which only provides an assessment of the edge particle recycling level. Two-dimensional distribution, as illustrated in Figure~\ref{fig:firstIMG} (d), of neutral particles is more effective for studying the behavior of neutral particles. 



With this task in mind, we find that Zhang et al.~\citep{zhang2025tomography} treat it as a linear inverse problem and solve a linear system of equations for the 2D distribution reconstruction. More specifically, they first pre-compute a weight matrix $\mathbf{W}$, which reflects the relationship between the camera line of sight and a discretized poloidal grid. Then, they employ the Simultaneous Adaptive Algebraic Reconstruction Technique (SAART) to improve emission estimation by comparing the reconstructed image with actual measurements. Lu et al.~\citep{lu2021tomographic} dedicate themselves to solving the problem of reconstructing the two-dimensional distribution of visible light radiation in the divertor region of the EAST tokamak. To solve this problem, the study employed the Philips-Tikhonov regularization method and introduced a second-order correction term to effectively suppress noise interference, thereby achieving high-precision and stable reconstruction of the radiation distribution on the poloidal section.

Although these methods have achieved promising results, they still suffer from the following limitations: 
\textbf{Firstly}, the problem is severely underdetermined, leading to an ill-posed inverse problem that is highly sensitive to noise and yields non-unique solutions. In addition, these methods are entirely based on physical principles and the solution process is fixed; they cannot utilize the powerful representation and optimization capabilities of data-driven neural networks. Therefore, in environments with strong noise, the solutions of these methods are correspondingly noisy or even erroneous. 
\textbf{Secondly}, the iterative reconstruction is computationally expensive and struggles to meet real-time performance requirements. As shown in Table~\ref{tab:efficiency}, even the fastest conventional method, e.g., the Simultaneous Adaptive Algebraic Reconstruction Technique (SAART)~\citep{lu2004adaptive,karhunen2019effect}, requires approximately 10 seconds to reconstruct a single image, which falls far short of real-time performance requirements. 
\textbf{Thirdly}, the handcrafted models/priors used in regularization are often insufficient to capture the rich structural complexity of the input images.
The above review and reflection motivate us to explore further \textit{how to design a more efficient and accurate deep neural network surrogate model to better address this problem.}

In this paper, we propose an efficient differential Transformer-based surrogate model to achieve fast and accurate 2D inverse distribution reconstruction, termed Delta-InvFormer. We formulate this 2D distribution reconstruction problem as directly regressing the intensity values at corresponding locations using a deep neural network, given the video frames. This end-to-end network mapping efficiently extracts visual features while avoiding the complex iterative optimization procedures of traditional models, resulting in higher computational efficiency. Moreover, a carefully designed differential Transformer module effectively suppresses interference from noise signals, thereby ensuring the accuracy of the final reconstruction. 
Specifically, given the input frames, we first partition and project them into visual tokens and capture the spatial-temporal features using a hybrid Transformer architecture, incorporating the standard Transformer, Spatial-Difference Transformer (S-DiffFormer), and Temporal-Difference Transformer (T-DiffFormer). This differential Transformer is built upon Flash Attention~\citep{dao2022flashattention} and effectively mitigates noise interference by modeling the differences between dual attention matrices~\citep{ye2024differential}. Finally, we adopt a decoder network and a projection layer to regress the 2D distribution maps. 



To sum up, the contributions of this work can be summarized into the following three main contributions: 

1). We propose an efficient and accurate Transformer-based surrogate model for visible CCD imaging diagnosis of neutral particle emission tomography in the EAST divertor, termed Delta-InvFormer. 
{It achieves a favorable accuracy-efficiency trade-off, providing competitive reconstruction accuracy while achieving approximately $\times 200$ faster inference speed.}

2). We propose S-DiffFormer and T-DiffFormer, two novel differential attention modules that model spatial and temporal feature differences, respectively, enabling robust and efficient 2D inversion under noisy conditions. 

3). Extensive experiments conducted on real data from the EAST divertor fully validated the effectiveness of our proposed surrogate model. More in detail,  
{compared with representative classical and recent deep learning baselines, Delta-InvFormer achieves the best overall performance across two test shots.}

\textit{The rest of this paper is organized as follows:} 
In Section~\ref{sec::relatedWorks}, we review the related works with a focus on the deep learning based nuclear fusion and Transformer networks. Then, we give a formal formulation of the research problem in Section~\ref{sec::problemFormulation}. After that, we introduce our proposed method for this 2D distribution problem in Section~\ref{sec::ourApproach}. The experiments are conducted on the real data from the EAST fusion device, and more details can be found in Section~\ref{sec::experiments}. Finally, we conclude this paper and point out the future works in Section~\ref{sec::conclusion}.

\section{Related Works} \label{sec::relatedWorks}
In this section, we introduce the related works on the fusion algorithms based on deep learning and Transformer networks. More details can be found in the following surveys~\citep{pavone2023MLFusionSurvey, barbarino2020Fusionbrief} and paper list~\footnote{\url{https://github.com/Event-AHU/AI4Fusion_Survey}}. 

{
\subsection{Deep Learning in Fusion}
In recent years, deep learning has been extensively employed in magnetic confinement fusion, encompassing parameter prediction, data-driven simulation, event prediction and identification, and temporal evolution modeling. 
For plasma state prediction and control-oriented modeling, Wang et al.~\citep{wang2024multi} used attention mechanisms to fuse 2D images and 1D diagnostic signals for Q-distribution prediction. 
Wu et al.~\citep{wu2024high} developed a data-driven tokamak dynamics model for reinforcement learning-based magnetic control, 
while Wai et al.~\citep{wai2022neural} proposed neural network models for plasma equilibrium and shape control in NSTX-U. 
For event identification and disruption prediction, Yang et al.~\citep{yang2025implementing} addressed limited data and distribution drift in disruption prediction through predict-first networks, data augmentation, and pseudo-data placeholder techniques. Ferreira et al.~\citep{ferreira2019deep} used CNN and LSTM models for plasma tomography and disruption prediction from bolometer data in JET. In addition, Gill et al.~\citep{gill2024real} proposed a 3D CNN-based confinement-regime classifier, while Shin et al.~\citep{shin2020real} and He et al.~\citep{he2024identifying} explored learning-based methods for L-H transition and ELM-related event identification.

Recently, fusion-oriented foundation models and large language models have also attracted attention. For example, Wang et al. proposed XiHeFusion~\citep{wang2025xiHeFusion} for science communication in nuclear fusion, and Yang et al.~\citep{yang2025fusionmae} introduced FusionMAE to learn unified plasma-state representations from diagnostic signals. 
More closely related to this work, machine learning has been explored for plasma reconstruction and tomographic inversion tasks. van Leeuwen et al.~\citep{van2025machine} proposed machine learning-based tomography reconstruction methods for real-time radiation tomography using multispectral imaging data on TCV. Dong et al.~\citep{dong2025adapted} proposed a real-time non-magnetic plasma shape reconstruction method from CCD images on HL-3. Wang et al.~\citep{wang2025physics} introduced a physics-informed deep learning model for line-integral diagnostics, and Zheng et al.~\citep{zheng2024real} developed a multi-task neural network for real-time magnetic equilibrium reconstruction on HL-3. These studies demonstrate that learning-based methods can effectively accelerate diagnostic reconstruction and provide real-time or near-real-time estimates of plasma states. 
}

{
\subsection{Transformer Network}
Transformer~\citep{vaswani2017attention} has been widely used for modeling long-range dependencies through self-attention and parallel matrix operations. In computer vision, Vision Transformer (ViT)~\citep{dosovitskiy2020image} divides an image into patch tokens and models their global relationships using Transformer blocks, showing strong visual representation ability. To improve efficiency and multi-scale feature extraction, Xie et al.~\citep{xie2021segformer} proposed SegFormer, which uses a hierarchical Mix Transformer backbone and a lightweight decoder for dense prediction tasks. These properties make Transformer-based architectures suitable for image-to-image regression and reconstruction problems. 
}

{
\subsection{Deep Learning in Tomographic Inversion}
In addition to the above-mentioned learning-based reconstruction studies in nuclear fusion diagnostics, similar inverse reconstruction problems have also been investigated in other tomographic imaging fields. Machine learning methods have been explored to learn data-driven priors or nonlinear inverse mappings for ill-posed reconstruction tasks. For example, Lei et al.~\citep{lei2024deep} proposed a deep algorithm-unrolling framework for electrical capacitance tomography, where physics-informed priors and adaptive regularization are learned to improve reconstruction quality. Dong et al.~\citep{dong2024acoustic} combined sparse reconstruction with a multiscale feature-extraction network for acoustic tomography temperature-field reconstruction. Lei et al.~\citep{lei2022physics} further introduced a physics-informed multi-fidelity learning method to improve the imaging quality of electrical capacitance tomography.

These studies indicate that learning-based methods can provide effective data-driven priors for ill-posed inverse reconstruction problems. 
Inspired by these works, we adopt a Transformer-based surrogate model for the EAST divertor neutral particle emission inversion task. Building on the efficient multi-scale representation of SegFormer, we further incorporate spatial and temporal differential modeling to improve reconstruction accuracy and robustness under noisy diagnostic conditions.
}

\section{Problem Formulation}  \label{sec::problemFormulation}

In the divertor plasma diagnostics of a tokamak device, the $\mathcal{D}_\alpha$ emission images acquired by a high-resolution tangential visible light camera system, as shown in  Figure~\ref{fig:firstIMG} (b), record the line integral radiation intensity of neutral particles in the divertor region. However, the emission images only provide a global view of the peripheral particle cycle and cannot directly reveal the two-dimensional poloidal distribution of neutral particle emission, which is crucial for understanding neutral particle transport and plasma-wall interaction processes. Therefore, the goal of this work is to recover the two-dimensional poloidal cross section of neutral particle emission in the divertor region from the $\mathcal{D}_\alpha$ camera images, which can be represented by a fixed $75 \times 98$ poloidal grid.

Classical tomographic approaches formulate this task as an ill–posed linear inverse problem. Let $\mathbf{E} \in \mathbb{R}^{N \times 1}$ denote the discretized poloidal emission distribution with $N =75 \times 98$, and let $\mathbf{W} \in \mathbb{R}^{HW \times N}$ be the weight matrix determined solely by the camera position and viewing geometry. The corresponding forward model from the poloidal emission to the camera image can be written as follows:
\begin{equation}
    \mathbf{S} = \texttt{Reshape}(\mathbf{W}\mathbf{E}),
    \label{eq:s}
\end{equation}
where $\mathbf{S} \in \mathbb{R}^{H \times W}$ denotes the measured $\mathcal{D}_\alpha$ image and $\texttt{Reshape}(\cdot)$ converts the vectorized image back to its two–dimensional layout. Traditional methods then reconstruct $\mathbf{E}$ from $\mathbf{S}$ by solving this linear system with appropriate regularization.

To gain a deeper understanding of this equation, we will provide a detailed explanation of its weight matrix $\mathbf{W}$ and discretized poloidal emission distribution matrix $\mathbf{E}$.

\noindent $\bullet$ \textbf{The Weight Matrix W:} The element $W_{ij}$ is the contribution of the $E_j$ to $S_i$, which is the $i$-th camera pixel. And different regions correspond to different $\mathbf{W}$ and $\mathbf{E}$ matrices.
For positions that are not visible to the camera, the corresponding positions of the $\mathbf{E}$ matrix are zero, and the corresponding columns of the W matrix are also zero. 

\begin{figure}
\centering 
\includegraphics[width=0.6\textwidth]{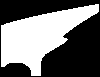}
\caption{Mask of the inversion matrix}
\label{fig:mask}
\end{figure}

\noindent $\bullet$ \textbf{The Inversion Matrix E:} 
The matrix can be obtained by solving Eq.~(\ref{eq:s}). 
It is worth noting that due to the camera's perspective, areas not visible to the camera cannot be inverted. 
Therefore, as shown in Figure~\ref{fig:mask}, we mask the areas that cannot be inverted.
Traditional methods for solving Eq.~(\ref{eq:s}) include Maximum  entropy method~\citep{denisova1998maximum}, Phillips-Tikhonov (P-T) regularization~\citep{lee2010modified}, and the Simultaneous  Adaptive Algebraic Reconstruction Technique (SAART)~\citep{lu2004adaptive,karhunen2019effect}. 
However, even the computationally fastest SAART method still requires 10 seconds to reconstruct a single image, making real-time implementation unfeasible.

\noindent $\bullet$ \textbf{The Surrogate Model:} Zhang et al.~\citep{zhang2025tomography} used the Simultaneous  Adaptive Algebraic Reconstruction Technique (SAART) to solve Eq.~(\ref{eq:s}). Although it has good performance and further accelerates the process, it still cannot meet the real-time requirements. Therefore, in order to speed up the calculation process while ensuring accuracy, we propose an efficient transformer-based surrogate model.

In this work, we instead treat divertor neutral emission reconstruction as a regressive problem and use {a neural network} model (i.e., the Delta-InvFormer) to approximate the inverse mapping of Eq.~\eqref{eq:s}. Concretely, the input to our model is the measured camera image $\mathbf{S} \in \mathbb{R}^{H \times W}$, and the desired output is the corresponding poloidal emission distribution $\mathbf{E} \in \mathbb{R}^{75 \times 98}$. A neural network parameterized by $\boldsymbol{\theta}$ is employed to learn this mapping
\begin{equation}
    \mathbf{E}_{out} = f_{\boldsymbol{\theta}}(\mathbf{S}),
    \label{eq:nn_mapping}
\end{equation}
where $f_{\boldsymbol{\theta}}(\cdot)$ serves as a nonlinear, data–driven inverse operator. 
During training, the network parameters $\boldsymbol{\theta}$ are optimized to minimize the discrepancy between the network prediction $\mathbf{E}_{out}$ and the reference emission distribution $\mathbf{E}$ by solving the following equation:
\begin{equation}
    \boldsymbol{\theta}^\star = \arg\min_{\boldsymbol{\theta}} \, \mathcal{L}\!\left(f_{\boldsymbol{\theta}}(\mathbf{S}), \mathbf{E}\right),
    \label{eq:learning_objective}
\end{equation}
where $\mathcal{L}(\cdot,\cdot)$ denotes the loss function. In this way, the neural network implicitly learns to invert the projection described in Eq.~\eqref{eq:s}, mapping $\mathcal{D}_\alpha$ images to the two-dimensional neutral particles emission distribution.

\begin{figure*}
\centering
\includegraphics[width=\textwidth]{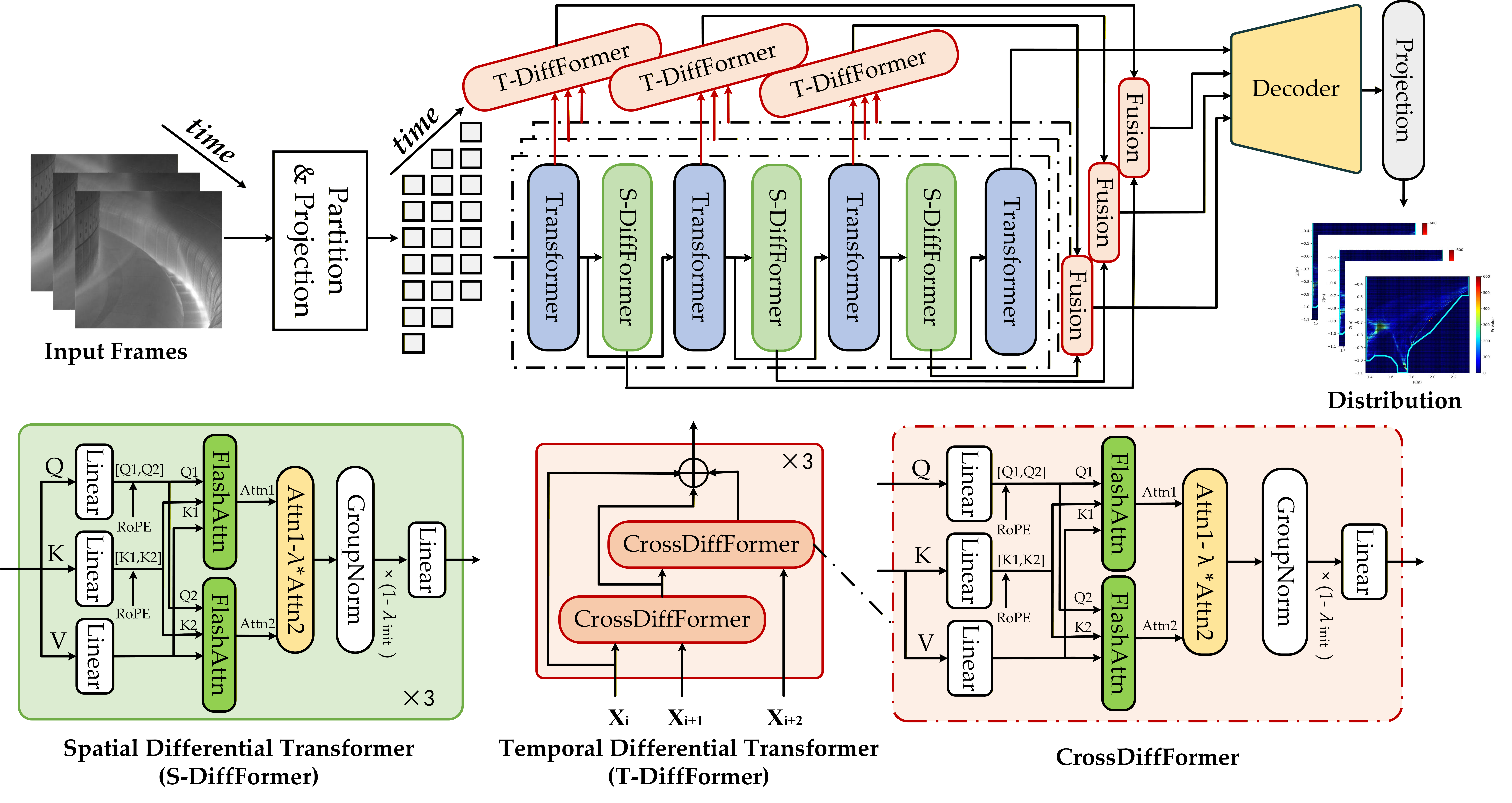}
\caption{An overview of our proposed Delta-InvFormer framework for efficient and accurate 2D inverse distribution reconstruction.}
\label{fig:framework}
\end{figure*}


\section{Our Proposed Approach} \label{sec::ourApproach}

In this section, we will first give an overview of our proposed Delta-InvFormer framework. Then, we will introduce the details on the input encoding, network architectures, and loss function used for the optimization.

\subsection{Overview} 
The overall architecture of our surrogate model is shown in Figure~\ref{fig:framework}. Specifically, we use continuous video frames as input to capture motion cues, then partition them into non-overlapping patches and project them into visual tokens. Then, a spatio-temporal backbone network is designed to encode these tokens based on the Differential Transformer Network~\citep{ye2024differential}, which contains the Spatial Differential Transformer (S-DiffFormer), the Temporal Differential Transformer (T-DiffFormer), and a decoder module. Note that the S-DiffFormer and T-DiffFormer are injected into a pre-trained { MiT} backbone network, which focuses on learning the spatial features of each frame and motion cues between different frames. The key operations of the two modules concentrate on suppressing the noisy information and achieving enhanced spatio-temporal feature learning. Finally, we aggregate these features using a fusion layer and feed them into the decoder network for distribution prediction.

\subsection{Input Encoding}  
Given the emission image sequence $\mathcal{D}_{\alpha} \in \mathbb{R}^{T \times H \times W \times 3}$, where $T$ is the number of frames in a video sample, $H, W$ denote the height and width of the frame, respectively, we first divide each frame into non-overlapping patches with the resolution $4 \times 4$. Then, a projection layer is adopted to convert these patches into visual tokens, which are then fed into a hierarchical transformer-based backbone network to obtain multi-scale features. Different from existing works, which usually add the position encoding to the token embeddings, inspired by the work~\citep{xie2021segformer}, we only feed the token embedding into the backbone network: 
\begin{equation}
    \mathbf{X} = \texttt{Encode}(\mathbf{I}),
    \label{eq:encode}
\end{equation}
where $\mathbf{I} \in \mathbb{R}^{T \times H \times W \times C}$ denotes the emission image sequence, $\mathbf{X} \in \mathbb{R}^{T \times \frac{H}{4} \times \frac{W}{4} \times D}$ is the visual tokens and $D$ is the feature dimension.

\subsection{Network Architecture} 
In this work, we adopt the Mix Transformer (MiT)~\citep{xie2021segformer} from SegFormer as the backbone network, due to the limited computational efficiency of ViT~\citep{dosovitskiy2020image}. However, traditional attention mechanisms focus excessively on irrelevant regions, and the backbone network serves only as an image feature extractor, lacking temporal information. Thus, we introduce a differential-Transformer-based network architecture that consists of an S-DiffFormer Block, a T-DiffFormer Block, a Fusion Module, a Decoder, and a Projection Layer. The first two model the features at the spatial and temporal levels, respectively, the Fusion Module is used to fuse spatial and temporal information, and the Decoder and Projection Layer complete the mapping from features to the final output.

\vspace{6pt}
\noindent \textbf{1) Spatial DiffFormer Block} 
\vspace{3pt}

Ye et al.~\citep{ye2024differential} originally introduced the Differential Transformer to NLP tasks for suppressing attention noise and amplifying signals that are truly relevant to the downstream objective. This was inspired by noise-canceling headphones and differential amplifiers in electrical engineering~\citep{laplante2018comprehensive}, which utilize the difference between two signals to eliminate common-mode noise. Inspired by its success in pruning redundant context, in this paper, we extend this philosophy to the visual domain and present S-DiffFormer, a spatial differential self-attention module that can be seamlessly plugged into any CNN or ViT backbone. Its input $\mathbf{X}_i \in \mathbb{R}^{N \times D}$ is the visual features extracted by the MiT backbone network, where $i$ represents the $i$-th frame in the input $\mathbf{X} \in \mathbb{R}^{T \times N \times D}$, where $T$ is the number of consecutive input frames, $N$ refers to the number of visual tokens, and $D$ denotes the embedding dimension of each token. Then, these visual tokens are fed into the S-DiffFormer module to enhance the original features extracted by the backbone network. This process can be formulated as follows: 
\begin{equation}
    \begin{aligned}
    \mathbf{Q}=[\mathbf{Q}_1,\mathbf{Q}_2] &= \texttt{Split}(\mathbf{X}_i\mathbf{W}_Q), \\
    \mathbf{K}=[\mathbf{K}_1,\mathbf{K}_2] &= \texttt{Split}(\mathbf{X}_i\mathbf{W}_K), \\
    \end{aligned}
    \label{eq:QK}
\end{equation}
\begin{equation}
    \mathbf{V}=\mathbf{X}_i\mathbf{W}_V,
    \label{eq:V}
\end{equation}
where $\mathbf{Q}$, $\mathbf{K}$, and $\mathbf{V}$ represent the query, key, and value, respectively. These are obtained by performing a linear transformation on the input features, and $\mathbf{W}_Q, \mathbf{W}_K, \mathbf{W}_V \in \mathbb{R}^{D \times D}$ are the learnable parameters in the neural network. It is worth noting that $\mathbf{Q}$ and $\mathbf{K}$ have been augmented with RoPE to enhance position awareness before attention calculation.

Next, based on the similarity between the query and the key, two attention matrices $\mathbf{Attn}_1$ and $\mathbf{Attn}_2$ are calculated. Then, according to the principle of differential noise suppression in digital signal processing, the weighted difference between the two attentions is calculated. The calculation process is as follows:
\begin{equation}
    \mathbf{Attn}_1 = \texttt{softmax}(\frac{\mathbf{Q}_1\mathbf{K}_1^T}{\sqrt{d_k}}),
    \label{eq:attn1}
\end{equation}
\begin{equation}
    \mathbf{Attn_2} = \texttt{softmax}(\frac{\mathbf{Q}_2\mathbf{K}_2^T}{\sqrt{d_k}}),
    \label{eq:attn2}
\end{equation}
\begin{equation}
    \mathbf{DiffAttn} = \mathbf{Attn}_1 - \lambda \cdot \mathbf{Attn}_2,
    \label{eq:diffattn}
\end{equation}
\begin{equation}
    \lambda=\exp{(\mathbf{y}_{q_1} \cdot \mathbf{y}_{k_1})}-\exp{(\mathbf{y}_{q_2} \cdot  \mathbf{y}_{k_2})} + \lambda_{init},
    \label{eq:lambda}
\end{equation}
where $\mathbf{y}_{q_1}$, $\mathbf{y}_{q_2}$, $\mathbf{y}_{k_1}$ and $\mathbf{y}_{k_2}$ are learnable vectors, $\lambda_{init} \in (0,1) $ is a constant for the initialization of $\lambda$.
Furthermore, due to the use of a multi-head mechanism, we employ the $GroupNorm(\cdot)$ operation for normalization, and finally, concatenate the multi-head {outputs} with the following equations:
\begin{equation}
    \mathbf{Z} = (1-\lambda_{init})\cdot \texttt{GroupNorm}(\mathbf{DiffAttn}),
    \label{eq:GropuNorm}
\end{equation}
\begin{equation}
    \mathbf{X}_S = \mathbf{Z}\mathbf{W}_O,
    \label{eq:GropuNorm}
\end{equation}
where $\mathbf{W}_O$ denotes the projection layer parameters.

\vspace{6pt}
\noindent \textbf{2) Temporal DiffFormer Block} 
\vspace{3pt}

As the input frames are continuous, we think that the $D_{\alpha}$ emission image not only affects the 2D distribution at the current moment, but also affects the distribution at the next moment. During the motion of neutral particles, the tokamak device is static, and the device background can be considered as noise across multiple frames. Therefore, we introduce a Temporal DiffFormer block, called T-DiffFormer, based on differential transformers. On the one hand, thanks to the static nature of the device, the background noise is common-mode across different frames, thus, we use differential transformers to reduce the device's influence on the inversion. On the other hand, to capture temporal dependencies, we use a cross-attention mechanism to capture the temporal dependencies between multiple frames. Therefore, we propose a Temporal DiffFormer Block to learn the temporal dependency. The structure of T-DiffFormer is shown in Figure~\ref{fig:framework} below, centered. 
The core of T-DiffFormer is CrossDiffFormer which is illustrated in Figure~\ref{fig:framework} to capture the temporal dependencies between consecutive frames. We use the previous frame $\mathbf{X}_i \in \mathbb{R}^{N \times D}$ as the query and $\mathbf{X}_{i+1} \in \mathbb{R}^{N \times D}$ as the key and value to capture the temporal dependency $\mathbf{T}_{i+1} \in \mathbb{R}^{N \times D}$ between two frames. 
Based on this, we use $\mathbf{T}_{i+1}$ as the query and $\mathbf{X}_{i+2} \in \mathbb{R}^{N \times D}$ as the key and value to capture its temporal dependency $\mathbf{T}_{i+2} \in \mathbb{R}^{N \times D}$. We then concatenate $\mathbf{X}_{i}$, $\mathbf{T}_{i+1}$, and $\mathbf{T}_{i+2}$ for subsequent fusion of temporal and spatial features, which can be written as: 
\begin{equation}
    \mathbf{T}_{i+1}=\texttt{CrossDiffFormer}(\mathbf{X}_i,\mathbf{X}_{i+1}),
    \label{eq:cdf1}
\end{equation}
\begin{equation}
    \mathbf{T}_{i+2}=\texttt{CrossDiffFormer}(\mathbf{T}_{i+1},\mathbf{X}_{i+2}),
    \label{eq:cdf2}
\end{equation}
\begin{equation}
    \mathbf{X}_{T}=\texttt{Concat}(\mathbf{X}_i,\mathbf{T}_{i+1},\mathbf{T}_{i+2}),
    \label{eq:concat}
\end{equation}
where $\mathbf{X}_{T} \in \mathbb{R}^{T \times N \times D}$ is the temporal feature.

\vspace{6pt}
\noindent \textbf{3) Fusion Block} 
\vspace{3pt} 

As shown in Figure~\ref{fig:framework}, we design a Fusion Module to fuse spatial features $\mathbf{X}_{S}$ and temporal features $\mathbf{X}_{T}$. 
First, we use two $1 \times 1$ Conv2d layers to downsample the $\mathbf{X}_{S}$ and $\mathbf{X}_{T}$ in the feature dimension, and then concatenate the downsampled $\mathbf{X}_{S}$ and $\mathbf{X}_{T}$ in the feature dimension, i.e., 
\begin{equation}
    \mathbf{\hat{X}}=\texttt{Concat}(\texttt{Conv2d}(\mathbf{X}_{S}), \texttt{Conv2d}(\mathbf{X}_{T})),
    \label{eq:fusion}
\end{equation}
where $\mathbf{\hat{X}}$ is the fusion feature.

\vspace{6pt}
\noindent \textbf{4) Decoder Network}  
\vspace{3pt} 

The baseline model SegFormer contains a lightweight decoder consisting only of MLP layers and upsampling layers. The decoding process first uses MLP layers and upsampling to align multi-scale features to a common resolution. These processes can be described as follows: 
\begin{equation}
    \begin{split}
        &\mathbf{F}_i = \texttt{Linear}(C_i,C)(\mathbf{\hat{X}}_i),~~i \in \{1, 2, 3\} \\ 
        &\mathbf{\tilde{F}}_i = \texttt{Upsample}(\frac{H}{4},\frac{W}{4})(\mathbf{F}_i),~~i \in \{1, 2, 3\} \\
        &\mathbf{F}_4 = \texttt{Linear}(C_4,C)(\mathbf{X}_4), \\
        &\mathbf{\tilde{F}}_4=\texttt{Upsample}(\frac{H}{4},\frac{W}{4})(\mathbf{F}_4), \\
    \end{split}
    \label{eq:decoder1}
\end{equation}
where $\mathbf{X}_4$ represents the features of the $4$-th layer transformer block of backbone, $\mathbf{\hat{X}}_i$, $i=\{1, 2, 3\}$ represents the feature obtained by fusing the $i$-th layer S-DiffFormer and T-DiffFormer features. These aligned features are then concatenated. Finally, a $1 \times 1$ Conv2d layer is applied to the aggregated features to generate the final prediction, i.e., 
\begin{equation}
\begin{split}
    &\mathbf{\tilde{X}}=\texttt{Conv2d}(\texttt{Concat}[\mathbf{\tilde{F}}_1,...,\mathbf{\tilde{F}}_4]),
    \label{eq:decoder2}
\end{split}
\end{equation}

\vspace{6pt}
\noindent \textbf{5) Projection Layer} 
\vspace{3pt} 

The projection layer is designed to transform the high-dimensional features learned by the network into the final 2D polar inversion image. Its architecture comprises two convolutional layers followed by two linear layers. This sequential structure progressively refines the feature representation to achieve the desired spatial output dimensions corresponding to the inversion grid. Since our input is only of the partially visible $D_{\alpha}$ emission image from the camera's limited sightlines, the output of the projection, inversion matrix $\mathbf{\hat{E}}$, contains a part that the camera cannot see in addition to the divertor part. As illustrated in Figure~\ref{fig:mask}, the white portion corresponding to the mask is the inversion region of the input image. To correct for the parts of the inversion matrix that are invisible to the camera, we apply the spatial mask to the output of Projection Layer, effectively setting the {unobservable} regions to zero and retaining the solution only for the observable divertor area. 
{In addition, we use the mean and standard deviation of $shot\#131076$ to perform Z-score normalization on the inversion matrix of the training data.}
The final prediction can be obtained via:
\begin{equation}
    \mathbf{\hat{E}} = \texttt{Projection}(\mathbf{\tilde{X}}), 
    \mathbf{E_{out}} = \texttt{MASK}(\mathbf{\texttt{Clamp}(\hat{E}}*\sigma+\mu,0,600)),
    \label{eq:projection}
\end{equation}
where $\mathbf{\hat{E}}$ is the output of Projection layer, $\mathbf{E_{out}}$ is the final 2D distribution. 
{We employ the operation $\mathrm{Clamp}(\cdot,0,600)$, which limits the reconstructed emission intensity to a physically reasonable range of $0$ to $600$ using \texttt{torch.clamp}.
This operation ensures the non-negativity of the reconstructed emission intensity, avoiding physically unreasonable negative values, while suppressing abnormally large predictions, thus improving the physical plausibility of the model output.
Finally, we further apply a spatial mask to remove regions not observed by the camera.}

\subsection{Loss Function}  
{In this paper, we adopt the Mean Squared Error (MSE) as the loss function for the optimization of our model.
The smaller the MSE value, the closer the model's prediction is to the reference result. 
Its calculation formula is as follows:}
\begin{equation}
    \mathcal{L}_{MSE} = \frac{1}{N}\sum_{i=1}^{N} (\mathbf{E}_i-\mathbf{\hat{E}}_i)^2,
    \label{eq:Dynamic}
\end{equation}
where $N =75 \times 98$ is the number of grids, 
{$\mathbf{E}$ is the ground truth,
$\mathbf{\hat{E}}$ is the prediction of our method.
$\mathbf{E}_i$ and $\mathbf{\hat{E}}_i$ represent the element of $\mathbf{E}$ and $\mathbf{\hat{E}}$, respectively, at the $i$-th grid. }


\section{Experiments} \label{sec::experiments}

\subsection{Dataset and Evaluation Metric}  
Our dataset comprises {$10145$} paired samples collected from {$9$} discharges (shots) on EAST (Experimental Advanced Superconducting Tokamak). It is the world's first fully superconducting tokamak fusion experimental device designed and built independently in China. It is located at the Institute of Plasma Physics, Chinese Academy of Sciences, Hefei, Anhui Province. More in detail, among the {$9$} shots, 7 shots ($shot\#131075$, $shot\#131076$, $shot\#131077$, $shot\#131078$, $shot\#131079$, $shot\#131080$, $shot\#131081$) are used for training, providing a total of $8793$ paired samples, {while the remaining 2 shots ($shot\#131082$ and $shot\#131083$) are reserved for testing, with {$1352$} paired samples. }
\begin{figure}[!htbp]
\centering
\includegraphics[width=\linewidth]{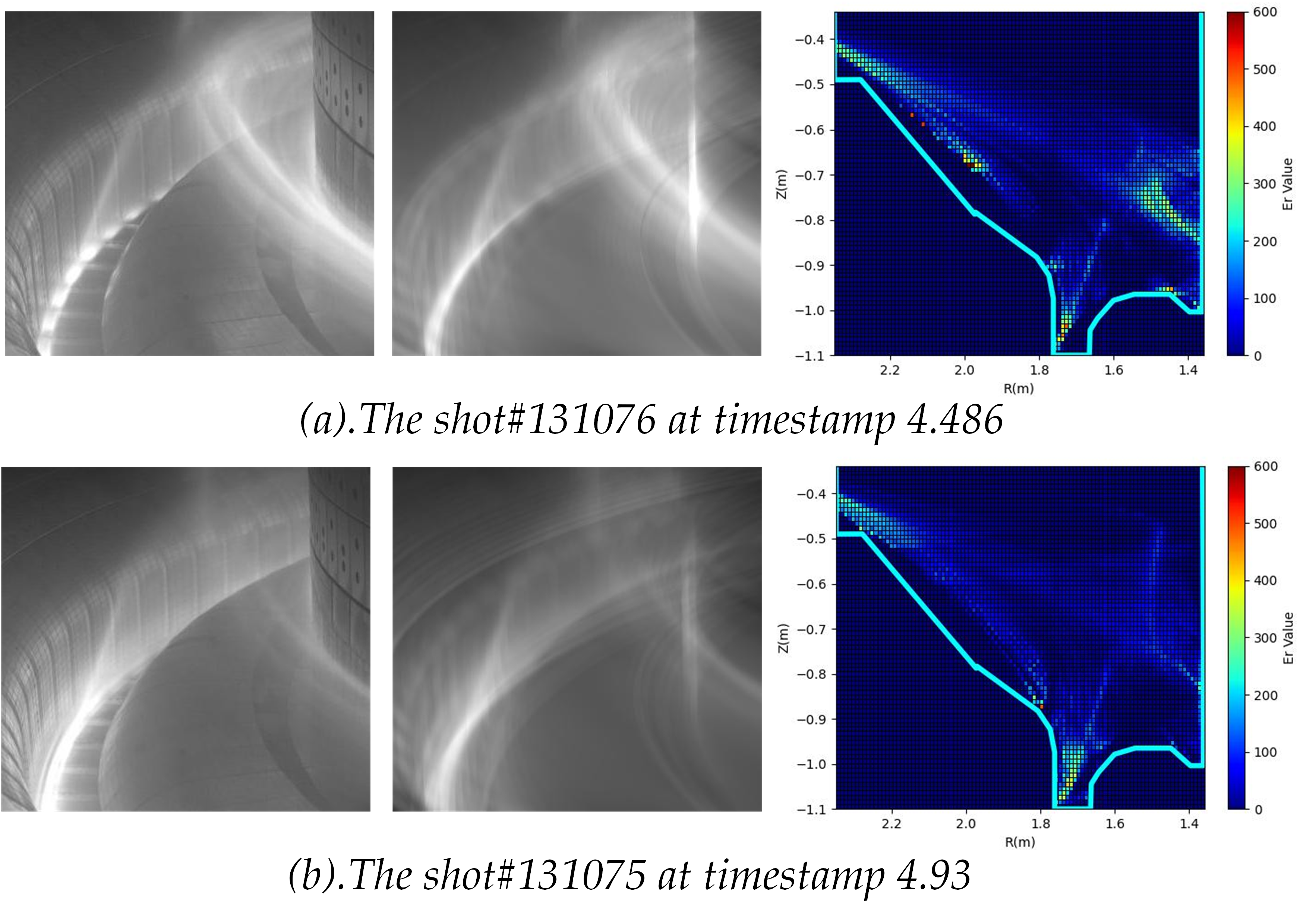}
\caption{Some representative shots in our dataset.} 
\label{fig:datasets}
\vspace{-6pt}
\end{figure}
The data have been carefully checked and verified by experienced scientists to ensure their physical reliability and usability. Each sample consists of (i) a measured tangential-view $\mathcal{D}_\alpha$ camera image, (ii) the corresponding two-dimensional poloidal cross-section distribution on a $75 \times 98$ grid obtained from traditional reconstruction methods, and (iii) a synthetic camera image generated by forward projecting this two-dimensional poloidal cross-section distribution back to the camera view. An overview of representative samples in our dataset is shown in Figure~\ref{fig:datasets}.

{In our experiments, the Mean Relative Error (MRE) is adopted as the primary evaluation metric to quantify the relative deviation between the reconstructed results and the reference results. MRE is a non-negative metric, and a smaller MRE indicates that the reconstructed result is closer to the reference result, corresponding to higher reconstruction accuracy. The formula is as follows:}
\begin{equation}
    \text{MRE}=\frac{1}{HW}\sum_{i=1}^{HW}{\left|\frac{\mathbf{S}_i-\hat{\mathbf{S}}_i}{\mathbf{S}_i}\right|},
    \label{eq:MRE}
\end{equation}
where $\mathbf{S}_i$ denotes the ground-truth value of the $i$-th pixel {in the camera image $\mathbf{S}$}, and $\hat{\mathbf{S}}_i$ denotes the corresponding reconstructed value {in the reconstructed image $\mathbf{\hat{S}}$}.

{
In addition, to directly evaluate the reconstructed two-dimensional $\mathcal{D}_\alpha$ emission distribution, we further adopt \textbf{MSE}, \textbf{MAE}, and \textbf{SSIM} as supplementary metrics. These metrics are computed on the predicted distribution and the corresponding reference distribution, providing a more direct assessment of the numerical error and structural similarity in the inversion domain.}

\subsection{Implementation Details} 
All experiments are conducted on a single NVIDIA GeForce RTX 3090 GPU with 24 GB of memory. We use the AdamW~\citep{loshchilov2017decoupled} optimizer and
the batch size is fixed to $1$, and the model is trained for $15$ epochs. 
Both the S-DiffFormer and T-DiffFormer blocks consist of three layers, which are inserted into layers 1, 2, and 3 of the backbone network. 
{Apart from our approach, all other pre-trained models are configured strictly according to their official default settings. 
Specifically, while Vision-LSTM does not load the pre-trained weights, it still adopts the architecture of the official base version. 
To adapt these models to our task, we only modify their output layers or prediction heads.

Further implementation details are shown in Table~\ref{tab:implementation}.}
Three consecutive raw frames are used as the input to our model. Before being fed into the network, each frame is resized to a spatial resolution of $512\times512$ pixels. The Mean Squared Error (MSE) is adopted as the loss function to guide optimization. 

\begin{table}[]
\centering
\caption{Implementation details and training settings of different models.}
\label{tab:implementation}
\resizebox{1.0\textwidth}{!}{
\begin{tabular}{c|cccccc}
\toprule
\toprule
\textbf{Model}  & \textbf{Pretrain}   & \textbf{Resolution}   & \textbf{LR Scratch}    & \textbf{LR Finetune}   & \textbf{Optimizer}  & \textbf{Batch Size} \\ 
\midrule

ResNet-50~\citep{he2016deep}        & ImageNet-1K  & $512 \times 512$  & 1e-4  & 5e-5 & AdamW  & 1 \\ \midrule

ResNet-101~\citep{he2016deep}       & ImageNet-1K  & $512 \times 512$  & 1e-4  & 5e-5 & AdamW  & 1  \\ \midrule

ViT-B~\citep{dosovitskiy2020image}            & ImageNet-1K  & $512 \times 512$  & 1e-4  & 5e-5 & AdamW  & 1   \\ \midrule

ViT-L~\citep{dosovitskiy2020image}             & ImageNet-1K  & $512 \times 512$  & 1e-4  & 5e-5 & AdamW  & 1   \\ \midrule

SwinT-B~\citep{liu2021swin}          & ImageNet-1K  & $512 \times 512$  & 1e-4  & 5e-5 & AdamW  & 1   \\ \midrule

SwinT-L~\citep{liu2021swin}          & ImageNet-1K  & $512 \times 512$  & 1e-4  & 5e-5 & AdamW  & 1   \\ \midrule

UNet~\citep{ronneberger2015u}            & ImageNet-1K  & $512 \times 512$  & -      & 5e-5 & AdamW  & 1   \\ \midrule


VMamba~\citep{liu2024vmamba}           & ImageNet-1K  & $512 \times 512$  & 1e-4  & 5e-5 & AdamW  & 1   \\ \midrule

ViL-B~\citep{alkin2025vision}            & -            & $512 \times 512$  & 1e-4  & -    & AdamW  & 1   \\ \midrule

UniRepLKNet-B~\citep{ding2024unireplknet}    & ImageNet-22K & $512 \times 512$  & 1e-4  & 5e-5 & AdamW  & 1   \\ \midrule

OverLoCK-B~\citep{lou2025overlock}       & ImageNet-1K  & $512 \times 512$  & 1e-4  & 5e-5 & AdamW  & 1   \\ \midrule

Delta-InvFormer  & Ade20K       & $512 \times 512$  & 1e-5  & 1e-4 & AdamW  & 1   \\ \bottomrule

\end{tabular}
}
\end{table}

\begin{figure*}[t]
\centering
\includegraphics[width=\linewidth]{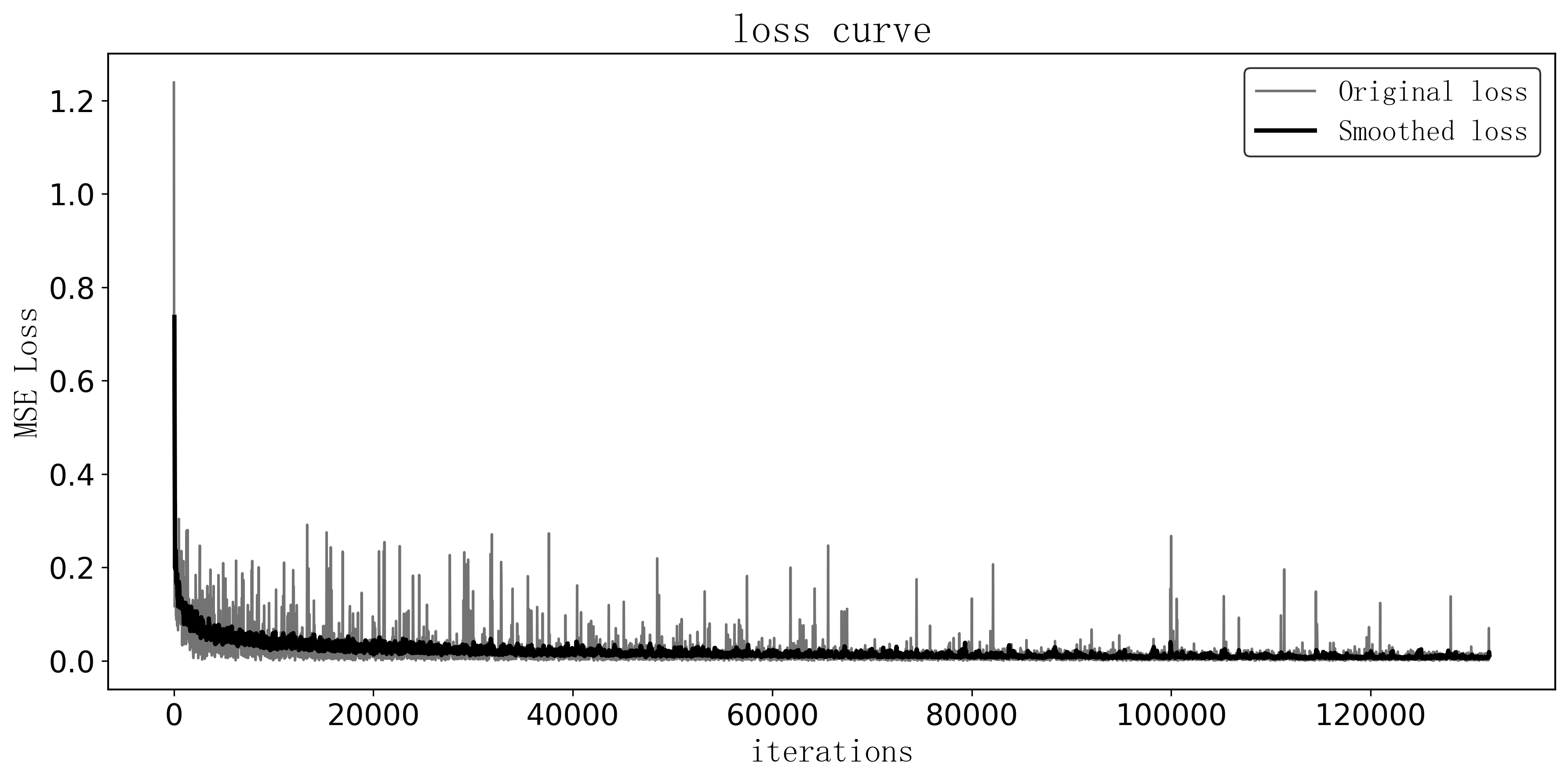}
\caption{Training convergence curve of Delta-InvFormer.} 
\label{fig:loss_curve}
\vspace{-6pt}
\end{figure*}

{
Furthermore, we present the training convergence curve in Figure ~\ref{fig:loss_curve}.
The training loss steadily decreases and converges during training. 
This indicates that, under the adopted training settings, the proposed model can be stably optimized without significant divergence or violent oscillations.
}

\begin{table}[h]
\centering
\caption{Quantitative comparison with other methods. \textbf{MRE} is computed on the reconstructed images, while \textbf{MSE}, \textbf{MAE}, and \textbf{SSIM} are computed on the predicted two-dimensional distribution (\textbf{E}).}
\label{table: comparision experiment new}
\setlength{\tabcolsep}{3pt}
\resizebox{1.0\textwidth}{!}{
\begin{tabular}{l|cc|cccccc}
\toprule
\toprule
\multirow{2}{*}{\textbf{Method}} & \multicolumn{2}{c|}{\multirow{2}{*}{\textbf{MRE$\downarrow$}}} & \multicolumn{6}{c}{\textbf{E}}\\
\cmidrule{4-9}
& \multicolumn{2}{c|}{} & \multicolumn{2}{c|}{\textbf{MSE$\downarrow$}} & \multicolumn{2}{c|}{\textbf{MAE$\downarrow$}} & \multicolumn{2}{c}{\textbf{SSIM$\uparrow$}} \\
\midrule
shot ID&\multicolumn{1}{c|}{131082}  &\multicolumn{1}{c|}{131083} &\multicolumn{1}{c|}{131082}  &\multicolumn{1}{c|}{131083} &\multicolumn{1}{c|}{131082}  &\multicolumn{1}{c|}{131083} &\multicolumn{1}{c|}{131082}  &\multicolumn{1}{c}{131083} \\
\midrule
ResNet50~\citep{he2016deep}&23.821 &25.059  &450.35  &790.54 &13.586 &15.444 &0.875 &0.857 \\
ResNet101~\citep{he2016deep} &30.351 &33.984 &489.026 &896.725 &14.678 &17.393 &0.864 &0.839 \\
ViT-B~\citep{dosovitskiy2020image} &6.515 &9.119 &108.472 &416.051 &5.627 &7.577 &\underline{0.971}  &0.957 \\
T-ViT-B~\citep{dosovitskiy2020image} & 15.585 &12.020 &271.982 &438.877 &9.772 &8.432 &0.926  &0.953 \\
ViT-L~\citep{dosovitskiy2020image} &\underline{6.069} &8.657 &110.668 &404.422 &5.660 &7.717 &0.967 &0.956 \\
SwinT-B~\citep{liu2021swin}        &8.145 &8.967 &\underline{95.871}  &\underline{262.809} &\underline{5.366} &\underline{5.993} &\underline{0.971} &\underline{0.972} \\
SwinT-L~\citep{liu2021swin}    &10.800 &13.264 &123.716  &324.607 &6.475 &7.478 &0.965 &0.962 \\
VMamba~\citep{liu2024vmamba}   &24.834 &19.988 &476.637  &544.436 &13.494 &11.401 &0.894 &0.923\\
UNet~\citep{ronneberger2015u}  &18.192  &14.614  &345.252	&509.664	&10.543	 &10.034	&0.9	&0.921   \\
T-UNet~\citep{ronneberger2015u} &25.186	&22.823	&552.369	&667.635	&14.272	&13.205	&0.857	&0.883   \\
ViL-B~\citep{alkin2025vision}   &6.913 &\textbf{6.424} &114.96 &333.685 &5.716 &6.478 &0.967 &0.967 \\
UniRepLKNet-B~\citep{ding2024unireplknet} &7.979 &\underline{6.452} &195.179 &323.488 &7.980 &7.506 &0.946 &0.954 \\
OverLoCK-B~\citep{lou2025overlock} &15.932 &23.726 &217.206 &395.821 &8.900 &9.783 &0.939 &0.943 \\
\midrule
Delta-InvFormer (Ours) &\textbf{4.638} &6.601 &\textbf{74.621} &\textbf{245.907} &\textbf{4.603} &\textbf{5.748} &\textbf{0.978} &\textbf{0.973} \\
\bottomrule

\end{tabular}
} 
\end{table}

\subsection{Comparison with Other Models} 
The problem investigated in this paper is a domain-specific task, for which no large-scale public dataset or a comprehensive suite of established baseline methods currently exists. 
This scarcity of readily available resources presents a significant challenge for comparative evaluation and benchmarking. 
{Therefore, to comprehensively evaluate the effectiveness of the proposed method, this paper retrains and compares several representative deep neural network models, including ResNet50~\citep{he2016deep}, ResNet101~\citep{he2016deep}, ViT-B~\citep{dosovitskiy2020image}, ViT-L~\citep{dosovitskiy2020image}, SwinT-B~\citep{liu2021swin}, SwinT-L~\citep{liu2021swin}, VMamba~\citep{liu2024vmamba}, UNet~\citep{ronneberger2015u} and the more recent ViL-B~\citep{alkin2025vision}, UniRepLKNet-B~\citep{ding2024unireplknet}, and OverLoCK-B~\citep{lou2025overlock}, which have strong visual modeling capabilities.
In addition, to verify the ability of ordinary backbone networks to utilize the temporal information of consecutive frames, we construct two temporal baseline models, T-ViT-B and T-UNet. 
Specifically, these two models use consecutive frames concatenated in the temporal dimension as input to evaluate whether simple multi-frame input is insufficient to capture inter-frame temporal relationships and improve performance.
All models are trained and tested on the same data partition and input resolution to ensure fairness in the comparison. 
The experimental results are shown in Table~\ref{table: comparision experiment new}.

To evaluate the reconstruction performance of different models, this paper reports results on two test discharge segments, $shot\#131082$ and $shot\#131083$, and uses MRE, MSE, MAE, and SSIM as evaluation metrics.
MRE measures the consistency of the model's predictions in the observed image space; while MSE, MAE, and SSIM are calculated on the two-dimensional emission distribution matrix and are used to directly assess the numerical error and structural similarity between the prediction and reference distributions.

As shown in Table~\ref{table: comparision experiment new}, traditional CNN models and some general-purpose visual backbone networks have relatively limited performance on this task.
For example, ResNet50~\citep{he2016deep}, ResNet101~\citep{he2016deep}, VMamba~\citep{liu2024vmamba}, and OverLoCK-B~\citep{lou2025overlock} have large reconstruction errors.
Transformer-based models perform better overall, with ViT-B~\citep{dosovitskiy2020image}, ViT-L~\citep{dosovitskiy2020image}, SwinT-B~\citep{liu2021swin}, and SwinT-L~\citep{liu2021swin} all significantly reducing MRE, indicating that the attention mechanism helps improve the consistency between the reconstructed image and the real observed image.

Further comparison of the higher-performing models reveals that ViL-B~\citep{alkin2025vision} and UniRepLKNet-B~\citep{ding2024unireplknet} achieve slightly lower MRE values ($6.424\%$ and $6.452\%$, respectively) on $shot \#131083$, while Delta-InvFormer achieves $6.601\%$.
However, Delta-InvFormer still achieves the best results on the predicted two-dimensional distribution ($\mathbf{E}$), with the lowest MSE and MAE of $245.907$ and $5.748$, respectively, and the highest SSIM of $0.973$ on $shot\#131083$.
Considering the results for $shot\#131082$, Delta-InvFormer remains the best overall.

Overall, Delta-InvFormer achieves the lowest average MRE on both test shots, while also obtaining the lowest average MSE, MAE, and highest average SSIM on the two-dimensional distribution ($\mathbf{E}$). 
This demonstrates that our method not only maintains good projection consistency in the reconstructed image space but also more accurately recovers the two-dimensional distribution of neutral particle radiation.
}



\newcommand{\yes}{\ding{51}}  
\newcommand{\no}{\ding{55}}   


\begin{table}[h]
\centering
\caption{Robustness comparison of different methods under Gaussian noise, uniform noise, and salt-and-pepper noise with different noise levels.} 
\label{table: noise}
\setlength{\tabcolsep}{3pt}
\resizebox{1.0\textwidth}{!}{
\begin{tabular}{c|r|rrr|rrr|rrr}
\toprule
\toprule
\multicolumn{1}{c|}{\multirow{2}{*}{\textbf{Model}}} 
& \textbf{\textit{w/o} Noise} 
& \multicolumn{3}{c|}{\textbf{Gaussian Noise}}  
& \multicolumn{3}{c|}{\textbf{Uniform Noise}} 
& \multicolumn{3}{c}{\textbf{Salt-and-Pepper Noise}} 
\\ \cline{2-11} 

\multicolumn{1}{c|}{}                       & \multicolumn{1}{c|}{-}         & \multicolumn{1}{c|}{3\%}    & \multicolumn{1}{c|}{5\%}    & \multicolumn{1}{c|}{10\%}    & \multicolumn{1}{c|}{3\%}    & \multicolumn{1}{c|}{5\%}    & \multicolumn{1}{c|}{10\%}   & \multicolumn{1}{c|}{3\%}    & \multicolumn{1}{c|}{5\%}     & \multicolumn{1}{c}{10\%}    \\ \midrule

ResNet50~\citep{he2016deep}  & 23.821 & 28.972 & 45.873 & 76.857  & 24.525 &27.685 & 52.951 &54.043 &70.010  & 98.702  \\ 

ResNet101~\citep{he2016deep} & 30.351  &77.031 &87.079 & 108.509 &59.18  &74.789 & 86.765 &96.974 &122.247 & 169.813 \\ 

ViT-B~\citep{dosovitskiy2020image}  & 6.515     &6.647  &\underline{8.214}  &\underline{26.475}  &6.484  &6.610   & \underline{9.837}  &28.321 &44.938  & 56.219  \\ 

ViT-L~\citep{dosovitskiy2020image}  &\underline{6.069}     &\underline{6.536}  &\textbf{7.343}  & \textbf{10.659}  &\underline{6.221}  &\underline{6.506}  &\textbf{7.981}  &\textbf{10.498} &\textbf{12.581}  &\textbf{14.767}  \\ 

SwinT-B~\citep{liu2021swin}  & 8.145     &27.644 &53.139 & 79.625  &16.278 &26.554 & 61.457 &54.294 &58.033  & 65.179  \\ 

SwinT-L~\citep{liu2021swin}  & 10.800      &24.579 &40.598 & 83.164  &18.02 &24.639 & 50.977 &26.543 &38.985  & 74.690   \\ 

VMamba~\citep{liu2024vmamba} & 24.834    &26.09  &28.247 & 56.93   &25.119 &26.019 & 30.464 &56.532 &80.000  & 122.125 \\ 

UniRepLKNet-B~\citep{ding2024unireplknet}  & 7.979     &55.03  &52.651 & 48.89   &40.012 &55.331 &51.479 &45.515 &44.834  & 44.673  \\ 

OverLoCK-B~\citep{lou2025overlock}  & 15.932    &37.242 &39.35  & 45.553  &35.293 &37.655 &40.371 &46.522 &46.376  & 46.964  \\ 

Delta-InvFormer (Ours)   &\textbf{4.638}     &\textbf{5.763}  &10.629 & 26.713  &\textbf{4.793}  &\textbf{5.527}  & 13.021 &\underline{28.100}   &\underline{33.358} &\underline{40.347}  \\ \bottomrule
\end{tabular}
}
\end{table}

{
\subsection{Robustness Analysis}
Considering potential interference during image acquisition and transmission, we artificially inject different levels of Gaussian noise, uniform noise, and salt-pepper noise ($3\%$, $5\%$, and $10\%$) into the input test frames to evaluate the model's robustness.

We present the robustness test results of the relevant models on $shot\#131082$, as shown in Table~\ref{table: noise}.
On the one hand, the reconstruction error (MRE) of most benchmark models increases with increasing noise intensity.
Under continuous noise interference (Gaussian and uniform noise), our method exhibits relatively stable performance.
For example, with $5\%$ uniform noise, the MRE of our model is $5.527\%$, outperforming CNN architectures of similar size and most visual Transformer models.
This indicates that our method can effectively suppress such continuous noise to a certain extent.

On the other hand, although ViT-L exhibits the least degradation when dealing with high-intensity discrete noise such as $10\%$ salt-pepper noise, its large computational overhead makes it difficult to meet the real-time requirements of control systems. 
In comparison, while Delta-InvFormer performs slightly worse under extreme noise, it achieves a better trade-off between accuracy and efficiency. 
It not only achieves the highest accuracy on raw data but also maintains reliable robustness in moderately noisy scenarios.
}

\begin{table}[h]
\centering
\caption{Uncertainty analysis under different random seeds on $shot\#131082$.}
\label{tab:uncertainty}
\resizebox{1.0\textwidth}{!}{
\begin{tabular}{c|cccc|cccc}
\toprule
\toprule
\textbf{Model}   & \multicolumn{4}{c|}{\textbf{ViT-B}~\citep{dosovitskiy2020image}}  & \multicolumn{4}{c}{\textbf{ViT-L}~\citep{dosovitskiy2020image}}  \\
\midrule
\textbf{Metrics}  & \multicolumn{1}{c}{\textbf{MRE$\downarrow$}} & \multicolumn{1}{c}{\textbf{MSE$\downarrow$}} & \multicolumn{1}{c}{\textbf{MAE$\downarrow$}} & \multicolumn{1}{c|}{\textbf{SSIM$\uparrow$}} & \multicolumn{1}{c}{\textbf{MRE$\downarrow$}}   & \multicolumn{1}{c}{\textbf{MSE$\downarrow$}}    & \multicolumn{1}{c}{\textbf{MAE$\downarrow$}}   & \multicolumn{1}{c}{\textbf{SSIM$\uparrow$}}  \\ 
\midrule
1234     & 6.515  & 108.472   & 5.627   &\underline{0.971}  &\underline{6.069}   & 110.668         & 5.66  & 0.967  \\

2345     & 6.856  & 122.857   & 5.987   & 0.968  &\underline{5.827}   & \underline{98.46}           &\underline{5.033}  &\underline{0.974}   \\

3456     & 8.264  & 136.579   & 6.334   & 0.964  &\textbf{6.326}  & 132.593          & 6.141  & 0.966    \\

mean & 7.212     & 122.636 & 5.983   & 0.968    &\underline{6.074}   & 113.907          & 5.611    & 0.969                     \\

std      & 0.757    & 11.476   & 0.289   & \underline{0.003}  &\underline{0.204}     & 14.122       & 0.454   & 0.004                     \\

\midrule
\textbf{Model}    & \multicolumn{4}{c|}{\textbf{SwinT-B}~\citep{liu2021swin}}  & \multicolumn{4}{c}{\textbf{Delta-InvFormer}}  \\
\midrule
\textbf{Metrics}  & \multicolumn{1}{c}{\textbf{MRE$\downarrow$}} & \multicolumn{1}{c}{\textbf{MSE$\downarrow$}} & \multicolumn{1}{c}{\textbf{MAE$\downarrow$}} & \multicolumn{1}{c|}{\textbf{SSIM$\uparrow$}} & \multicolumn{1}{c}{\textbf{MRE$\downarrow$}}   & \multicolumn{1}{c}{\textbf{MSE$\downarrow$}}    & \multicolumn{1}{c}{\textbf{MAE$\downarrow$}}   & \multicolumn{1}{c}{\textbf{SSIM$\uparrow$}}  \\ 
\midrule

1234  & 8.145  &\underline{95.871}  &\underline{5.366}  &\underline{0.971}  & \textbf{4.638}  &\textbf{74.621}  &\textbf{4.603} &\textbf{0.978} \\

2345  & 7.779  & 103.637 & 5.288  & 0.968  &\textbf{4.976}  &\textbf{85.916}  &\textbf{4.803} &\textbf{0.976} \\

3456  & 8.033  &\underline{104.209} &\underline{5.632}  &\underline{0.970}   &\underline{6.466}  &\textbf{73.76}   &\textbf{4.66}  &\textbf{0.976} \\

mean & 7.986   &\underline{101.239}   &\underline{5.429}   &\underline{0.970}  &\textbf{5.360}  &\textbf{78.099}  &\textbf{4.689}  &\textbf{0.977}                     \\

std      &\textbf{0.153}                   &\textbf{3.803}                   &\underline{0.147}                   &\textbf{0.001}                    & 0.794                     &\underline{5.539}                      &\textbf{0.084}                     &\textbf{0.001}  \\
\bottomrule
\end{tabular}
}
\end{table}

{
\subsection{Uncertainty Analysis}
To further evaluate the uncertainty of the model's prediction results under the influence of random factors, this paper selects ViT-B, ViT-L~\citep{dosovitskiy2020image}, and SwinT-B~\citep{liu2021swin} as representative comparative models and conducts uncertainty analysis with Delta-InvFormer.
Specifically, we train and test each model under different random seed settings, and calculate MRE, MSE, MAE, and SSIM, while also calculating the mean and standard deviation of each metric.
The results are shown in Table~\ref{tab:uncertainty}.

As can be seen from Table ~\ref{tab:uncertainty}, our method maintains good reconstruction performance under different random seeds.
Compared with ViT-B, ViT-L, and SwinT-B, our method achieves the lowest average MRE of $5.360\%$, outperforming ViT-B~\citep{dosovitskiy2020image} ($7.212\%$), ViT-L~\citep{dosovitskiy2020image} ($6.074\%$), and SwinT-B~\citep{liu2021swin} ($7.986\%$).
Meanwhile, Delta-InvFormer achieves the lowest average MSE and MAE ($78.099$ and $4.689$, respectively) on the two-dimensional distribution, and the highest average SSIM ($0.977$). 
This demonstrates that our proposed method not only maintains low error in the reconstructed image space but also recovers the two-dimensional distribution itself more accurately.

Standard deviation reflects the magnitude of performance fluctuations under different random seeds, i.e., the degree of uncertainty in the model's prediction results.
From the perspective of standard deviation, Delta-InvFormer's MAE and SSIM standard deviations are 0.084 and 0.001, respectively, both remaining at low levels, indicating that its results in the two-dimensional distribution reconstruction task have less fluctuation and lower uncertainty.
Although the standard deviation of Delta-InvFormer's MRE is not the lowest among all models, its average MRE is still significantly better than other comparative methods, indicating that this model can effectively control the performance fluctuations caused by random initialization and training processes while maintaining high reconstruction accuracy.

Overall, uncertainty analysis shows that Delta-InvFormer not only has excellent average reconstruction performance, but also maintains relatively stable prediction results under different random seed conditions, demonstrating good training stability and reliability under different random seeds.
}

\begin{table}[h]
\centering
\caption{Sensitivity analysis of Z-score normalization statistics.}
\label{tab:norm}
\resizebox{1.0\textwidth}{!}{
\begin{tabular}{c|cccc|cccc}
\toprule
\toprule
Shot ID  & \multicolumn{4}{c|}{131082}  & \multicolumn{4}{c}{131083}
\\ \midrule
Metrics  & \textbf{MRE}$\downarrow$  & \textbf{MSE}$\downarrow$   & \textbf{MAE}$\downarrow$   & \textbf{SSIM}$\uparrow$   & \textbf{MRE}$\downarrow$   & \textbf{MSE}$\downarrow$   & \textbf{MAE}$\downarrow$  & \textbf{SSIM}$\uparrow$  
\\ \midrule
131076   & \textbf{4.638} & \underline{74.621}  & \underline{4.603}  &\underline{0.978}    &\underline{6.601}  & 245.907   &5.748 & 0.973          
\\
131079  & \underline{4.88} & \textbf{74.417} & \textbf{4.602} &\textbf{0.979} & \textbf{6.45}  & \textbf{225.92}  & \textbf{5.343} & \textbf{0.976} 
\\
131080  & 19.883  & 98.393  & 7.077   & 0.965   & 22.753  & 250.852 &8.032  & 0.955 
\\
all  & 9.330 & 101.112 & 6.037  & 0.969  & 9.197  & 260.484  & 6.580 &0.971          
\\ \midrule
131076 $\xrightarrow{}$ 131079 & 10.257         & 126.743         & 6.494          & 0.964          & 9.046                         & 330.751                        & 7.318          & 0.955          
\\
131076$\xrightarrow{}$131080 & 10.836         & 102.158         & 6.496          & 0.954          & 15.465                        &301.157 & 8.303          & 0.938          
\\
131076$\xrightarrow{}$all    & 8.293          & 81.35           & 4.951          & 0.977          & 6.761                         & 251.05                         & \underline{5.604}          & \underline{ 0.974}    
\\ \midrule
131079$\xrightarrow{}$131076 & 13.124         & 135.807         & 6.951          & 0.97           & 12.024                        & 260.033                        & 6.878          & 0.972          
\\
131079$\xrightarrow{}$131080 & 18.806         & 106.295         & 6.988          & 0.966          & 20.457 & 243.538                        & 7.622          & 0.959          
\\
131079$\xrightarrow{}$all    & 9.339          & 113.207         & 6.229          & 0.969          & 7.972                         & \underline{ 240.281}                  & 6.294          & 0.972          
\\      \bottomrule    
\end{tabular}
}
\end{table}

{
\subsection{Sensitivity Analysis of Z-score Normalization Statistics}
To analyze the influence of normalization statistics, we conduct experiments using the mean and standard deviation computed from different training shots, as shown in Table~\ref{tab:norm}.
All statistics are calculated only within the valid reconstruction region, while invalid masked areas are excluded. 
Table~\ref{tab:norm} is divided into three parts. 

In the first part, the same normalization statistics are used for both training and testing. 
The results show that the statistics computed from different shots lead to different reconstruction performance.
For example, using the statistics of $shot\#131076$ achieves the best MRE on $shot\#131082$, while using the statistics of $shot\#131079$ obtains the best overall performance on $shot\#131083$. 
In contrast, the statistics of $shot\#131080$ lead to a clear performance degradation, indicating that not all shot-level statistics are equally suitable for normalization.

The second and third parts evaluate the cross-normalization setting, where `A$\rightarrow$B' means that the model is trained using the normalization statistics computed from $shot\#A$ and tested using the statistics computed from $shot\#B$.
These results show that directly changing the normalization statistics between training and testing can affect the reconstruction performance.
Further observation of the second and third parts of cross-normalization experiments reveals that model performance typically declines when standardized statistics from different sources are used during training and testing.
This indicates that mismatched standardized statistics alter the scale distribution of the input data and introduce additional distributional bias.
In contrast, in the first part of experiments, when both training and testing use the mean and standard deviation of $shot\#131076$, the model achieves the lowest MRE on $shot\#131082$ while maintaining good reconstruction performance on $shot\#131083$.
Although the statistics of $shot\#131079$ also have advantages on some metrics, $shot\#131076$ is more competitive in cross-standardization experiments, as shown in Table~\ref{tab:norm}, indicating that it provides a more suitable and stable standardized scale for the model.
Therefore, it is reasonable to use the mean and standard deviation calculated within the effective region of $shot\#131076$ as the standardization reference in this work.
}

{
\subsection{Training Stability Analysis with Gradient Accumulation}
Due to GPU memory limitations, the physical batch size in our experiments is set to 1. However, such a small batch size may affect training stability. 
Therefore, we further conduct gradient accumulation experiments to increase the effective batch size and examine whether the reported results are sensitive to small batch training settings.

As shown in Table~\ref{tab:accumulation}, Delta-InvFormer maintains the best average performance under the gradient accumulation setting, achieving an average MRE of $4.890\%$, MSE of $79.631$, MAE of $4.855$, and SSIM of $0.975$. 
Together with the uncertainty analysis in Table~\ref{tab:uncertainty}, these results indicate that the proposed method remains reliable under different random seeds and a larger effective batch size.
}

\begin{table}[h]
\centering
\caption{Effect of gradient accumulation on training stability.}
\label{tab:accumulation}
\resizebox{1.0\textwidth}{!}{
\begin{tabular}{c|cccc|cccc}
\toprule
\toprule
\textbf{Model}   & \multicolumn{4}{c|}{\textbf{ViT-B}~\citep{dosovitskiy2020image}}  & \multicolumn{4}{c}{\textbf{ViT-L}~\citep{dosovitskiy2020image}}  \\
\midrule
\textbf{Metrics}  & \multicolumn{1}{c}{\textbf{MRE$\downarrow$}} & \multicolumn{1}{c}{\textbf{MSE$\downarrow$}} & \multicolumn{1}{c}{\textbf{MAE$\downarrow$}} & \multicolumn{1}{c|}{\textbf{SSIM$\uparrow$}} & \multicolumn{1}{c}{\textbf{MRE$\downarrow$}}   & \multicolumn{1}{c}{\textbf{MSE$\downarrow$}}    & \multicolumn{1}{c}{\textbf{MAE$\downarrow$}}   & \multicolumn{1}{c}{\textbf{SSIM$\uparrow$}}  \\ 
\midrule
1234 & 8.848 & 133.779 & 6.576 & 0.96  & 10.52                        & 191.339 & 7.804 & 0.949 \\
2345 & 8.215 & 142.131 & 6.669 & 0.962 & 9.975                        & 142.484 & 6.707 & 0.961 \\
3456 & 7.308 & 137.573 & 6.493 & 0.963 & 9.208                        & 154.96  & 6.739 & 0.961 \\
mean     & 8.124 & 137.828 & 6.579 & 0.962 & 9.901                        & 162.928 & 7.083 & 0.957 \\
std      & 0.774 & 4.182   & 0.088 & 0.002 & 0.659                        & 25.383  & 0.624 & 0.007 \\
\midrule
\textbf{Model}    & \multicolumn{4}{c|}{\textbf{SwinT-B}~\citep{liu2021swin}}  & \multicolumn{4}{c}{\textbf{Delta-InvFormer}}  \\               
\midrule
\textbf{Metrics}  & \multicolumn{1}{c}{\textbf{MRE$\downarrow$}} & \multicolumn{1}{c}{\textbf{MSE$\downarrow$}} & \multicolumn{1}{c}{\textbf{MAE$\downarrow$}} & \multicolumn{1}{c|}{\textbf{SSIM$\uparrow$}} & \multicolumn{1}{c}{\textbf{MRE$\downarrow$}}   & \multicolumn{1}{c}{\textbf{MSE$\downarrow$}}    & \multicolumn{1}{c}{\textbf{MAE$\downarrow$}}   & \multicolumn{1}{c}{\textbf{SSIM$\uparrow$}}  \\ 
\midrule
1234 & 8.005 & 100.578 & 5.617 & 0.971 & 4.142                        & 70.812  & 4.491 & 0.978 \\
2345 & 9.897 & 104.087 & 6.075 & 0.969 & 5.433 & 84.268               & 5.092 & 0.974 \\
3456 & 8.5   & 92.405  & 5.382 & 0.973 & 5.095                        & 83.814  & 4.983 & 0.974 \\
mean     & 8.801 & 99.023  & 5.691 & 0.971 & 4.890                        & 79.631  & 4.855 & 0.975 \\
std      & 0.981 & 5.994   & 0.352 & 0.002 & 0.669                        & 7.641   & 0.320 & 0.002 \\
\bottomrule
\end{tabular}
}
\end{table}

\begin{table}[h]
\centering
\caption{The results of S-DiffFormer and T-DiffFormer with different layer settings.}
\label{table:layer}
\begin{tabular}{l|ccc|cll}
\toprule
\toprule
\textbf{Module} & \multicolumn{3}{c|}{\textbf{S-DiffFormer}} & \multicolumn{3}{c}{\textbf{T-DiffFormer}} \\ \midrule
Layer  & \multicolumn{1}{c|}{1}  & \multicolumn{1}{c|}{2}  & 3 & \multicolumn{1}{c|}{1} & \multicolumn{1}{c|}{2} & \multicolumn{1}{c}{3} \\ 
\midrule
MRE    & \multicolumn{1}{c|}{5.257\%} & \multicolumn{1}{c|}{6.538\%} & 5.139\% & \multicolumn{1}{c|}{6.795\%} & \multicolumn{1}{l|}{7.216\%} & 5.315\% \\ 
\bottomrule
\end{tabular}
\vspace{-6pt}
\end{table}

\subsection{Layer Study}
As shown in Table~\ref{table:layer}, we conduct ablation studies on the number of S-DiffFormer and T-DiffFormer layers to evaluate how the depth of each module affects the reconstruction performance. To isolate the contribution of each module, we vary one type of block at a time while removing the other; that is, when analyzing the number of S-DiffFormer layers, no T-DiffFormer blocks are inserted into the network, and vice versa. In these layer-number experiments, all S-DiffFormer (or T-DiffFormer) blocks are inserted into the 2nd, 3rd, and 4th layers of the backbone network. The results show that using three layers yields the best performance for both S-DiffFormer and T-DiffFormer. 

\begin{table}[h]
\centering
\caption{The results of selecting layers at different positions.}
\label{table: position}
\tabcolsep=0.4cm
\begin{tabular}{l|c|c|c|c}
\toprule
\toprule
\textbf{Position}       & \textbf{1-2-3}       & \textbf{1-2-4}       & \textbf{1-3-4}    & \textbf{2-3-4} \\ \midrule
\textbf{MRE}       & 4.638\%       & 5.855\%       & 5.092\%    & 5.735\%\\ \bottomrule
\end{tabular}
\vspace{-6pt}
\end{table}

\subsection{Position Study}
In this work, the backbone network extracts four layers of multi-scale features. Considering the real-time requirements of the control system, it is impractical to refine all of them. To balance efficiency and accuracy, we only select three out of the four stages for refinement using our module. The numbers of layers in S-DiffFormer and T-DiffFormer are both set to three.

To determine which backbone stages should be refined, we index the four stages from shallow to deep as $1$–$4$ and evaluate all combinations of three stages. As shown in Table~\ref{table: position}, using the first three stages ($1$–$2$–$3$) yields the lowest MRE of $4.638\%$. Replacing any of them with the deepest stage (configurations $1$–$2$–$4$, $1$–$3$–$4$, and $2$–$3$–$4$) leads to higher errors of $5.855\%$, $5.092\%$, and $5.735\%$, respectively. These results suggest that incorporating dense low- and mid-level features is more beneficial for accurate reconstruction than relying on the deepest high-level feature map, and therefore we adopt the $1$–$2$–$3$ configuration as the default in all subsequent experiments.

\subsection{Ablation Study}
As shown in Table~\ref{table: ablation study}, we conduct ablation studies on main modules to demonstrate the effectiveness of components. Specifically, the experiment ID $\#01$ represents the baseline model SegFormer~\citep{xie2021segformer}, $\#02$ represents the removal of the S-DiffFormer module, and $\#03$ represents the removal of the T-DiffFormer module.

\begin{table}[h]
\centering
\caption{Ablation Study of Module.}
\label{table: ablation study}
\tabcolsep=0.22cm
\resizebox{1.0\textwidth}{!}{
\begin{tabular}{c|cccccccc}
\toprule
\toprule
\textbf{Index} & \textbf{Baseline}  & \textbf{T-DiffFormer}  & \textbf{S-DiffFormer}  & \textbf{1234}   & \textbf{2345}  & \textbf{3456} &\textbf{Mean} &\textbf{Std}\\ \midrule
$\#01$ &  \yes    & \no    & \no   & 9.989    &8.387   &8.197  & 8.858 & \underline{0.804}  \\
$\#02$ &  \yes    & \yes    & \no  & 7.612    &\underline{5.139}   &\underline{6.749}  &\underline{6.263} &1.069\\ 
$\#03$ &  \yes    & \no   & \yes   & \underline{4.781}    &7.260   &7.873  & 6.875 &1.232\\ 
$\#04$ &  \yes    & \yes   & \yes  & \textbf{4.638}    &\textbf{4.976}   &\textbf{6.466}   & \textbf{5.360} &\textbf{0.794}\\
\bottomrule
\end{tabular}
}
\end{table}

\noindent $\bullet$ \textbf{Analysis on Baseline.~}
In this paper, we adopt the SegFormer~\citep{xie2021segformer} as our baseline and use the MiT backbone of SegFormer as the feature extractor to evaluate its performance on image reconstruction. As shown in Table~\ref{table: ablation study}, the SegFormer baseline achieves competitive MRE compared with the methods listed in Table~\ref{table: comparision experiment new}, and performs better than many CNN-based counterparts. These results suggest that MiT-based architectures are highly suitable and competitive for image reconstruction tasks.

\noindent $\bullet$ \textbf{Analysis on T-DiffFormer.~} 
As our baseline only captures the spatial features of $D_\alpha$ emission images and does not consider the temporal relationship between the input frames, as shown in Figure~\ref{fig:framework}, we design a Temporal DiffFormer Block based on the Differential Transformer and cross-attention mechanism to learn the temporal dependency. 
{
From Table~\ref{table: ablation study}, the average MRE drops from $8.858\%$ to $6.263\%$, indicating that the T-DiffFormer block also has excellent performance compared to the baseline. }

\noindent $\bullet$ \textbf{Analysis on S-DiffFormer.~} 
Although the traditional attention mechanism has excellent feature extraction capabilities, it still pays too much attention to irrelevant context, thereby weakening the attention to relevant context. Differential Transformer, proposed by Ye et al.~\citep{ye2024differential}, can amplify attention to the relevant context while canceling noise. Thus, we design a Spatial DiffFormer Block, illustrated in Figure~\ref{fig:framework} with three layers Differential Transformer to further extract spatial features. 
{
According to Table~\ref{table: ablation study}, compared with the baseline, the S-DiffFormer block achieves superior performance with average MRE decreasing from $8.858\%$ to $6.875\%$. }

{
Compared with the baseline model, both T-DiffFormer and S-DiffFormer bring clear performance improvements. Specifically, adding T-DiffFormer reduces the average MRE from $8.858\%$ to $6.263\%$, corresponding to a relative reduction of $29.3\%$. Adding S-DiffFormer alone reduces the average MRE to $6.875\%$, corresponding to a relative reduction of $22.4\%$. When both modules are used together, the average MRE further decreases to $5.360\%$, achieving a relative reduction of $39.5\%$ compared with the baseline. 
These results demonstrate that both temporal and spatial differential modeling are effective, and their combination provides complementary benefits for improving reconstruction accuracy.
}



\begin{table}[h]
\centering
\caption{Comparison of the FPS, Params, GFLOPs, Memory Usage and MRE on $shot\#131082$ and $shot\#131083$. Where {Mem.} denotes the Memory Usage}
\label{tab:efficiency}
\resizebox{1.0\textwidth}{!}{
\begin{tabular}{c|crccrr}
\toprule
\toprule
\textbf{Method} &\textbf{FPS} & \textbf{Params}  & \textbf{GFLOPs}  &\textbf{Mem.} & \textbf{131082} & \textbf{131083} \\
\midrule
P-T regularization~\citep{lee2010modified}  & 0.001  &-  &-  &\textgreater{}100(CPU) &-  &-            \\
SAART~\citep{zhang2025tomography}             & 0.1    &-  &-  &5.600    &\textbf{3.139}   &\textbf{5.562}             \\
ViT-B~\citep{dosovitskiy2020image}             &\textbf{30}     &\underline{92.023}   & 263.241  &\textbf{5.400} & 6.515  & 9.119    \\
ViT-L~\citep{dosovitskiy2020image}             & 13     & 311.861  & 931.678  & 6.340 &6.069  & 8.657    \\
SwinT-B~\citep{liu2021swin}           &\underline{22}     &94.392   &263.123  & \underline{5.548} & 8.145  & 8.967    \\
SwinT-L~\citep{liu2021swin}           & 16     & 206.465  & 591.436  & 6.302 & 10.800 & 13.264   \\
ViL-B~\citep{alkin2025vision}             & 14     & 98.067   & 265.484  & 5.800 & 6.913  &\underline{6.424}    \\
UniRepLKNet-B~\citep{ding2024unireplknet}     & 21     & 105.61   &\underline{255.442}  & 5.868 & 7.979  & 6.452    \\
OverLoCK-B~\citep{lou2025overlock}        & 13     & 103.000  & 276.814  & 5.696 & 15.932 & 23.726   \\
Delta-InvFormer   & 20     & \textbf{86.787}  & \textbf{229.212}  & 6.968 &\underline{4.638}  & 6.601     \\
\bottomrule
\end{tabular}
}
\end{table}
{
\subsection{Efficiency Analysis} 
In this work, we evaluate the performance of traditional inversion methods, representative deep learning models, and Delta-InvFormer on metrics such as FPS, parameter count, GFLOPs, memory usage (Mem.), and MRE.
The results are shown in Table ~\ref{tab:efficiency}. 
Mem. is in GB. 
For deep learning models and SAART, Mem. represents GPU memory usage during inference; for P-T regularization running on the CPU, it represents CPU memory usage.

As shown in Table~\ref{tab:efficiency}, the computational efficiency of traditional inversion methods is significantly limited. 
The fastest, SAART~\citep{zhang2025tomography}, only achieves $0.1$ FPS, which is still insufficient for fast inversion. 
In contrast, Delta-InvFormer achieves $20$ FPS, an improvement of approximately $200$ times compared to SAART, indicating that deep learning surrogate models can significantly accelerate the inversion process.

In comparisons among deep learning models, Delta-InvFormer achieves the lowest MRE on $shot\#131082$ and remains highly competitive on $shot\#131083$. 
Compared to some models with higher computational complexity, it exhibits smaller reconstruction errors; 
simultaneously, compared to ViT-L~\citep{dosovitskiy2020image} with its larger parameter count, Delta-InvFormer offers faster inference speed and higher inversion accuracy.

In conclusion, {Delta-InvFormer achieves a favorable balance between reconstruction accuracy and inference efficiency.}
Although its computational cost is not the lowest, it achieves a good balance between FPS, memory usage, and MRE, indicating that our proposed method is more suitable for EAST divertor $\mathcal{D}_\alpha$ emission inversion tasks where both accuracy and efficiency are required.

} 


\subsection{Visualization} 

\begin{figure*}[t]
\centering
\includegraphics[width=1\linewidth]{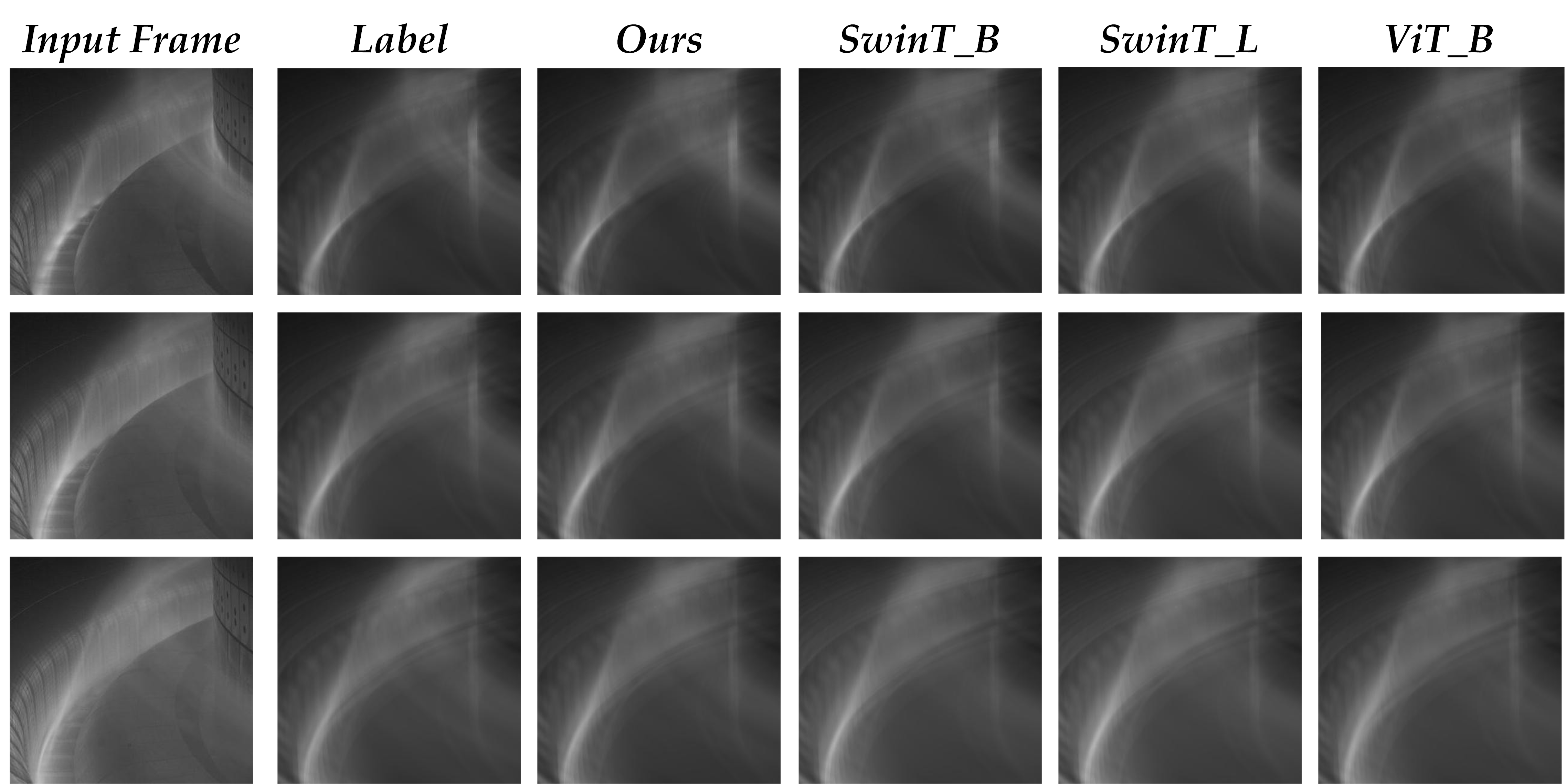}
\caption{Comparison of results for reconstructing the raw image from 2D brightness distribution using different neural networks.}   
\label{fig:reconstructed}
\end{figure*}

\begin{figure*}
\centering
\includegraphics[width=1\linewidth]{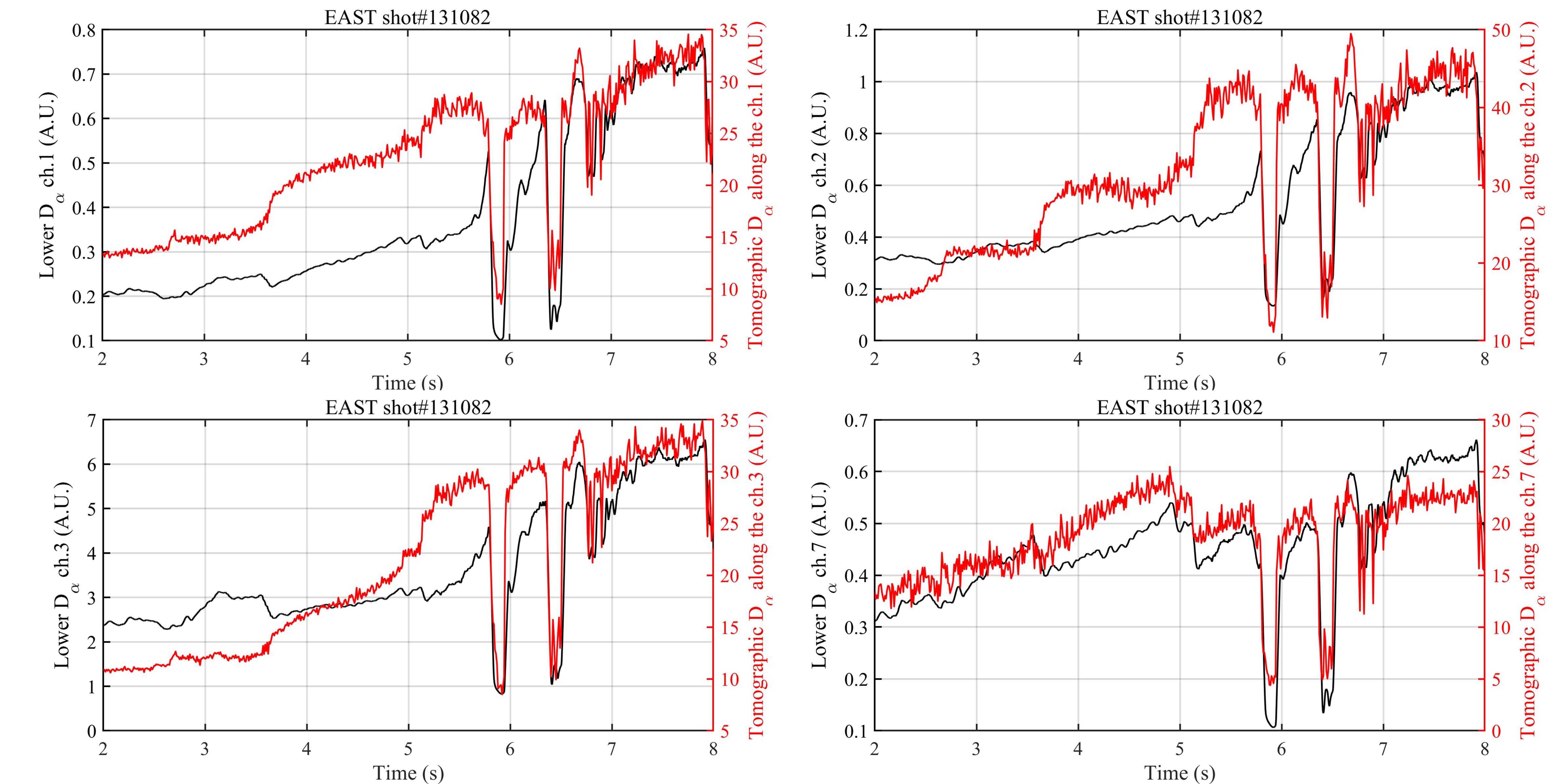}
\caption{Verification of the Delta-InvFormer reconstructed $\mathcal{D}_\alpha$ emission with the Filterscope measurement for EAST shot $\#$131082.}
\label{fig:placeholder}
\end{figure*}

As shown in Figure~\ref{fig:reconstructed},  we reconstruct the image using the W matrix and the E matrix and visualize it. The label is the real reconstructed image, and the others are the images reconstructed by our method and {the selected comparison methods.} 
The result of our method is the closest to the label image. 
Furthermore, to further verify the accuracy and reliability of Delta-InvFormer in the $\mathcal{D}_\alpha$ tomographic reconstruction of the EAST divertor region, we perform line integrals of the two-dimensional $\mathcal{D}_\alpha$ emission obtained based on Delta-InvFormer along the lines 
of sight corresponding to the {$1_{st}, 2_{nd}, 3_{rd}$, and $7_{th}$} channels of the Filterscope, and compare them with the measured $\mathcal{D}_\alpha$ signal of the lower divertor, as shown in Figure~\ref{fig:placeholder}.
As can be seen from Figure~\ref{fig:placeholder}, {the temporal evolution trends of the reconstructed and measured signals are generally consistent}: during the period of $2\text{–}5s$, the signal gradually increases, indicating enhanced $\mathcal{D}_\alpha$  radiation in the divertor region; 
{around the transient decrease at $5.8\text{--}6.0s$, we further conduct a quantitative analysis. For the measured Filterscope signals, the intensities of channels $1$, $2$, $3$, and $7$ decrease by approximately $80\%$, $79\%$, $78\%$, and $75\%$, respectively. The corresponding line-integrated signals obtained from the Delta-InvFormer reconstructed two-dimensional $\mathcal{D}_\alpha$ emission also show similar decreases, with reduction ratios of approximately $68\%$, $76\%$, $69\%$, and $72\%$ for channels $1$, $2$, $3$, and $7$, respectively. 
\begin{figure}[!htp]
\centering
\includegraphics[width=1.0\textwidth]{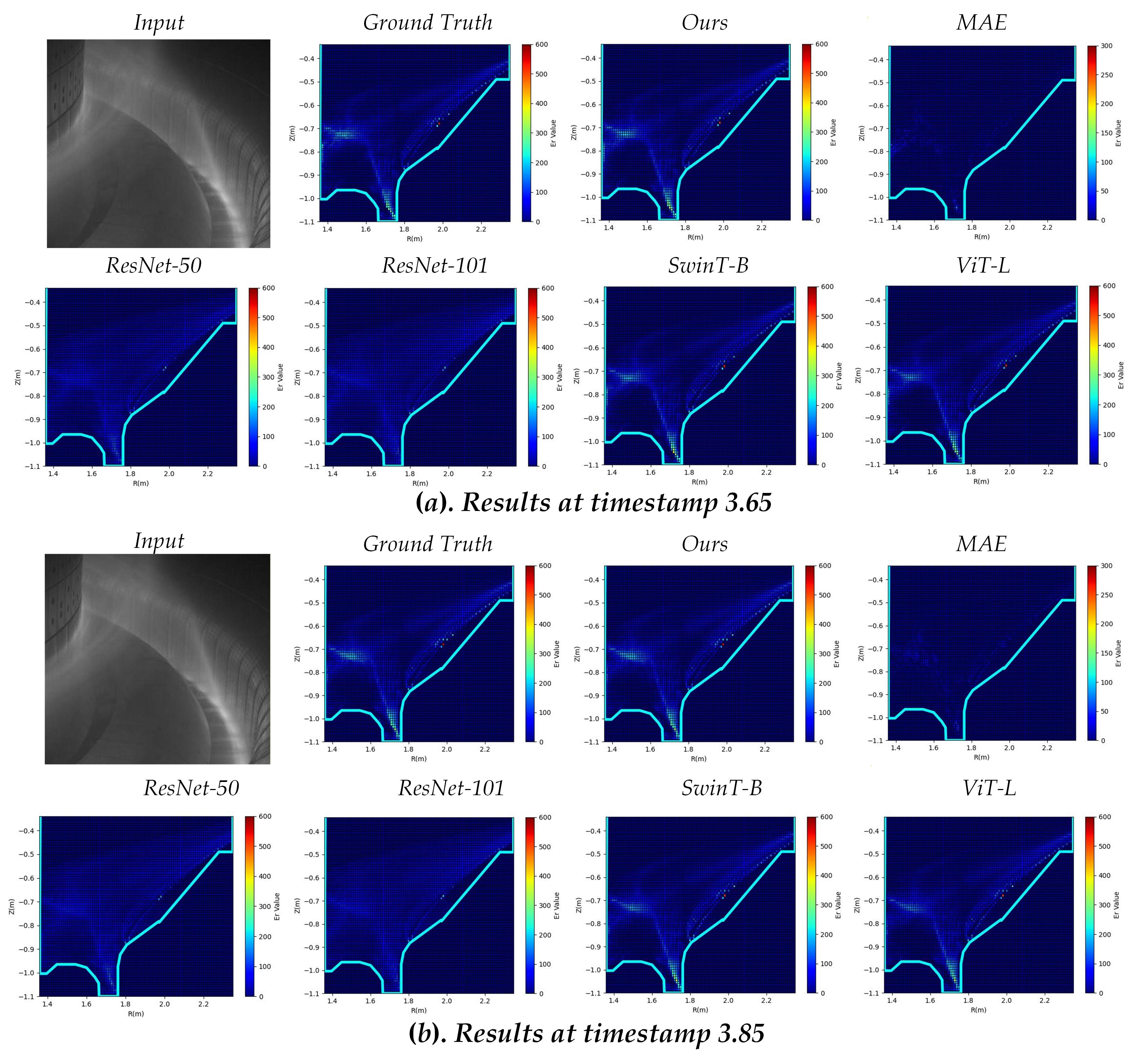}
\caption{Comparison of 2D brightness distribution prediction at different time steps between the proposed model and other strong neural networks.}
\label{fig:enter-label}
\vspace{-12pt}
\end{figure}
These results indicate that the reconstructed emission not only follows the overall temporal evolution trend, but also captures the short-time weakening and subsequent recovery of divertor $\mathcal{D}_\alpha$ radiation around $5.8\text{--}6.0s$.
Additionally, as shown in Table~\ref{tab:coefficient}, the Pearson correlation coefficients of the four channels are all above 0.82, indicating a strong positive correlation between the measured Filterscope signals and the line-integrated reconstructed signals. 
This further supports the temporal consistency of the reconstructed $\mathcal{D}_\alpha$ emission.
}

\begin{table}[h]
\centering
\caption{Pearson correlation coefficients between the measured Filterscope signals and the line-integrated reconstructed signals.} 
\label{tab:coefficient}
\begin{tabular}{c|cccc}
\toprule
\toprule
\textbf{Channel} &ch.1 &ch.2 &ch.3 &ch.7 \\
\midrule
\textbf{Pearson correlation coefficient} &0.863  &0.829  &0.851  &0.847 \\
\bottomrule
\end{tabular}
\vspace{-12pt}
\end{table}

Overall, the tomographic reconstruction results show good consistency with the Filterscope measurements, indicating that {this tomographic method can reasonably capture the main temporal evolution} of the divertor $\mathcal{D}_\alpha$ emission.

We visualize the results of our method and other comparative methods in Figure~\ref{fig:enter-label}. We select the time points of $3.65s$ and $3.85s$ to show our results, where Ground Truth is the real two-dimensional distribution, MAE is the difference between Ground Truth and our method, and others are the results of the other comparative models. The Transformer-based methods are clearly superior to the CNN-based method. The result of our method is close to the true distribution.

\subsection{Limitation Analysis} 
Although our proposed method significantly improves inversion efficiency while maintaining accuracy, it still has several limitations. 
First, limited by the camera's field of view on the EAST device, we can only acquire images of the region near the lower divertor. 
Therefore, the inversion results primarily reflect the particle emission intensity in the lower divertor region, failing to provide a complete global description of the core plasma. 
This spatial limitation may also lead to bias in the training data, as the model has never encountered emission modes associated with the upper divertor or other poloidal regions, thus its generalization ability remains unclear. 
Second, the current model is trained and evaluated on a relatively small dataset. 
This means the network may not fully capture variations in experimental conditions, and there is room for improvement in its generalization ability to unknown scenarios.
{In addition, since the supervised labels are generated based on traditional inversion methods, the reconstruction accuracy of Delta-InvFormer is also partly constrained by the accuracy and uncertainty of these reference results.}
Particularly, the training data may not adequately represent rare events or atypical emission modes, potentially leading to performance degradation when applied to more complex experimental conditions.
In the future, we will consider pre-training on a larger dataset, expanding the inversion area, and further accelerating the process. 
{
Finally, the current Delta-InvFormer provides deterministic point predictions and does not explicitly quantify prediction-level uncertainty. For future deployment in control-oriented or safety-critical fusion diagnostic scenarios, estimating the confidence of each reconstruction and identifying potentially unreliable predictions are important.
Therefore, future work will explore uncertainty quantification methods, such as MC-Dropout~\citep{gal2016dropout}, deep ensembles~\citep{lakshminarayanan2017simple}, and conformal prediction, to further improve the reliability of the surrogate model.}

\section{Conclusion} \label{sec::conclusion} 
This paper focuses on the 2D inverse reconstruction problem in fusion research and proposes a novel surrogate model based on a differential formulation, 
{significantly improving both the reconstruction efficiency and accuracy of this task.}
More specifically, we utilize a multi-scale transformer as the backbone network to extract raw image features, then use the S-DiffFormer module to further refine the extracted multi-scale features. To leverage the temporal dependencies between video frames, we introduce the T-DiffFormer module to learn the temporal dependencies between input frames, and use a fusion module to fuse spatial and temporal features. This allows our method to accelerate the reconstruction process while maintaining accuracy. 
{While the proposed method significantly improves reconstruction efficiency, it still falls short of meeting the real-time application requirements of CCD imaging systems under high sampling rates. Furthermore, the currently used dataset is relatively limited in size, which restricts the model's generalization ability across different discharge scenarios. Meanwhile, the EAST device has accumulated a large amount of one-dimensional diagnostic data with important physical information, but these data have not been fully utilized in this paper. Future work will further expand the training data scale, enabling the model to learn and adapt to evolutionary characteristics under different discharge conditions; simultaneously, we will continue to explore more efficient model acceleration schemes and perform multi-source information fusion in conjunction with one-dimensional diagnostic data to further improve the model's reconstruction accuracy and practical application capabilities.}

\section*{Acknowledgment} \label{sec::Acknowledgment} 
The authors acknowledge the High-Performance Computing Platform of Anhui University for providing computing resources. The authors also acknowledge the support and contributions of the EAST Team in Hefei (\url{https://cstr.cn/31130.02.EAST}) for providing technical support and assistance in data collection and analysis.


\bibliographystyle{apalike}
\bibliography{reference}


\end{document}